\documentclass{article} 
\PassOptionsToPackage{dvipsnames,table,xcdraw}{xcolor}
\usepackage[final]{colm2026_conference}

\usepackage{microtype}
\usepackage{hyperref}
\usepackage{url}
\usepackage{booktabs}

\usepackage{lineno}

\definecolor{darkblue}{rgb}{0, 0, 0.5}
\hypersetup{colorlinks=true, citecolor=darkblue, linkcolor=darkblue, urlcolor=darkblue}

\usepackage{colortbl}
\usepackage{wrapfig}

\usepackage{amsmath}
\usepackage{amssymb}
\usepackage{amsfonts}
\usepackage{amsthm}
\usepackage[mathcal]{eucal}
\usepackage{mathrsfs}
\usepackage{bm}

\usepackage{cleveref}

\usepackage[export]{adjustbox}
\usepackage{multirow}

\usepackage{algorithmic}

\usepackage[T1]{fontenc}
\usepackage[utf8]{inputenc}

\usepackage{graphicx}
\usepackage{subcaption}

\usepackage{caption}
\usepackage{placeins}
\usepackage{float}
\usepackage{needspace}

\usepackage{amsmath,amsfonts,bm}

\def\eqref#1{equation~\ref{#1}}

\def\1{\bm{1}}

\DeclareMathAlphabet{\mathsfit}{\encodingdefault}{\sfdefault}{m}{sl}
\SetMathAlphabet{\mathsfit}{bold}{\encodingdefault}{\sfdefault}{bx}{n}

\title{PreMoE: Proactive Inference for Efficient Mixture-of-Experts}

\author{Zehua Pei$^{1,*}$, 
Ying Zhang$^{2,*}$, 
Hui-Ling Zhen$^2$, 
Tao Yuan$^2$, 
Xianzhi Yu$^2$,\\
\bf Zhenhua Dong$^2$,
Sinno Jialin Pan$^1$,
Mingxuan Yuan$^2$, 
Bei Yu$^1$\\
$^1$The Chinese University of Hong Kong\\
$^2$Huawei Technologies Co., Ltd\\
$^*$Equal contribution.
}

\begin{document}

\ifcolmsubmission
\linenumbers
\fi

\maketitle

\begin{abstract} 
Mixture-of-Experts (MoE) models offer dynamic computation, but are typically deployed as static full-capacity models, missing opportunities for deployment-specific specialization.
We introduce PreMoE, a training-free framework that proactively compiles sparse MoE variants for targeted deployment scenarios.
At its core is Predicted Expert Utility (PEU), a robust metric for estimating expert importance from router logits through high-confidence threshold filtering and logit transformation, which together stabilize utility estimation under aggressive sparsity.
Using PEU scores computed on a small calibration set, PreMoE produces domain-aware expert rankings that can be used to compile either domain-specific specialists or high-efficiency multi-domain generalists, without any retraining.
Across MoE models ranging from 30B to 718B parameters, PreMoE preserves near-full accuracy at model-dependent sparsity levels of up to 50\%; on DeepSeek-R1, 50\% sparsity halves the deployment NPU count and increases throughput by 23\%.
Controlled studies across calibration sources, formats, random subsets, and domain mixtures confirm that these gains are stable rather than tied to a particular calibration set.
PreMoE further exposes a practical deployment trade-off: specialists maximize in-domain efficiency, while synthesized generalists retain broader cross-domain capability at the same sparsity budget. 
\footnote{Code: \url{https://github.com/JarvisPei/PreMoE}}

\end{abstract}

\section{Introduction}

Mixture of Experts (MoE) architectures~\citep{shazeer2017outrageously,lepikhin2020gshard,zhou2022mixture} have enabled unprecedented scaling of language models~\citep{jiang2024mixtral,dai2024deepseekmoe,guo2025deepseek,tang2025pangu}, yet despite their dynamic design, they face a practical contradiction: they are deployed statically.
This approach requires the entire set of experts to reside in memory, failing to capitalize on the model's potential for specialization and leading to significant inefficiency.
Prior work has explored expert pruning to address this challenge~\citep{chen2022task,muzio2024seer,lu2024not}, but these methods rely on coarse statistics like activation frequency that fail to capture activation quality, often requiring costly finetuning to recover performance, or employ search procedures that become intractable at scale.
We propose a proactive compilation approach that uses a refined signal to compile specialized MoE instances before deployment, without finetuning or expensive search.

\begin{figure*}[htbp]
    \centering
    \includegraphics[width=0.85\linewidth]{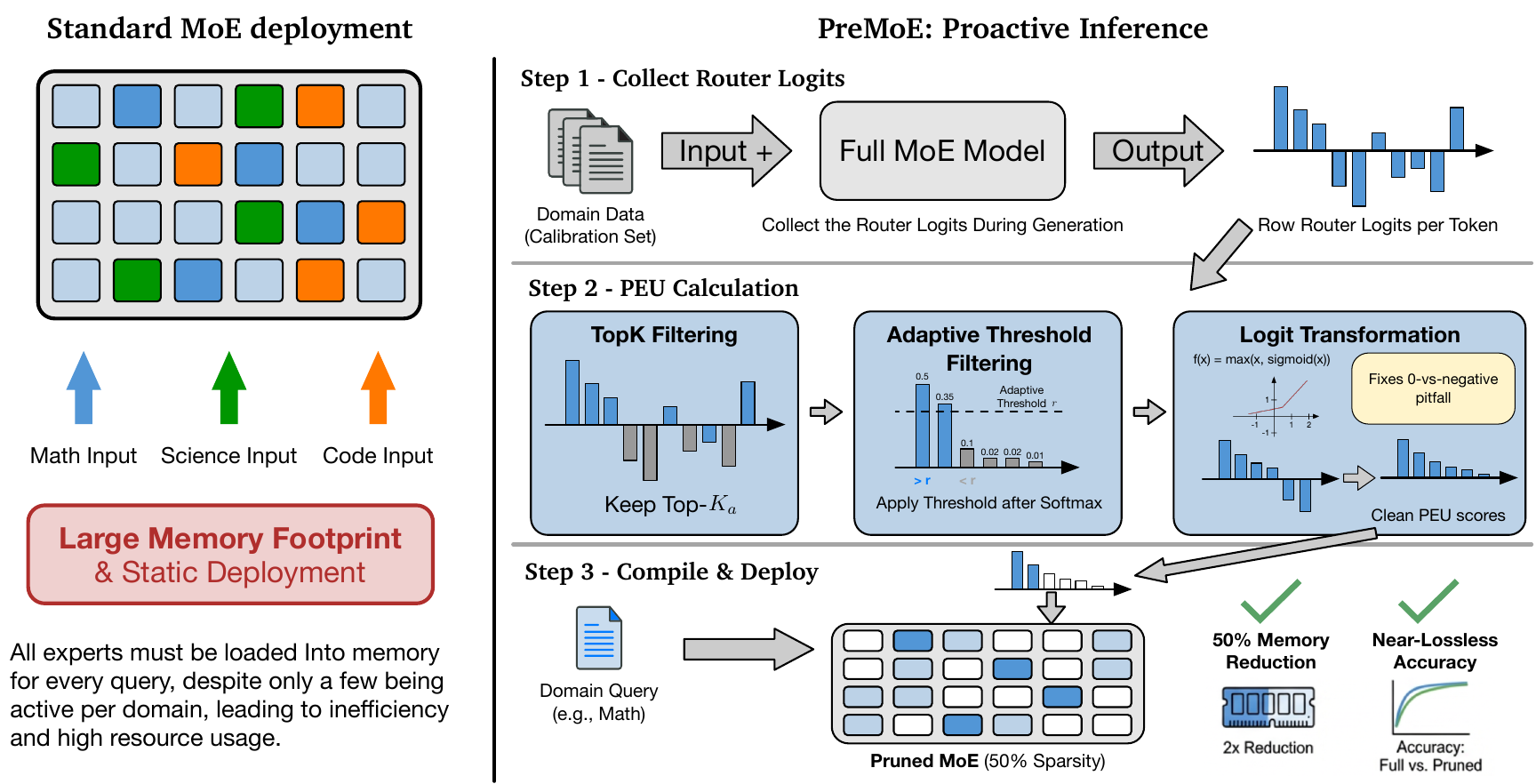}
    \caption{Overview of PreMoE. \textbf{Left:} Standard MoE deployment requires all experts in memory despite only a few being active per domain. \textbf{Right:} PreMoE's proactive inference pipeline: (1) collect router logits during generation on calibration data, (2) compute PEU scores via a processing pipeline (TopK filtering $\rightarrow$ Adaptive threshold filtering $\rightarrow$ Logit transformation), and (3) compile a pruned MoE with 50\% sparsity achieving near-lossless accuracy and 2$\times$ memory reduction.}
    \label{fig:premoe-overview}
    \vspace{-4mm}
\end{figure*}

We introduce \textbf{PreMoE}, a training-free framework that actualizes this approach (\Cref{fig:premoe-overview}).
At its heart is our novel Predicted Expert Utility (PEU) metric, which robustly measures expert importance by analyzing the model's native router logits through high-confidence filtering (TopK filtering and adaptive threshold filtering) and logit transformation.
By computing PEU scores for all experts across all layers on a domain-specific calibration dataset, PreMoE yields a layer-wise PEU ranking that quantifies which experts are critical for that domain.
We refer to this ranking as the domain's ``computational pattern'' and use it as a blueprint to identify minimal, high-performance expert subsets and compile specialized model instances.

\Needspace{9\baselineskip}
Our contributions are:
\vspace{-2mm}
\begin{itemize}
    \item We find that expert utility is highly domain-specific and predictable when measured with a principled signal (\Cref{fig:moe_logit_distribution}), enabling effective pre-deployment expert selection.
    \vspace{-1mm}
    \item We propose PreMoE, a training-free framework built on our Predicted Expert Utility (PEU) metric, which refines router logits via high-confidence filtering and logit transformation to compile specialized MoE instances.
    \vspace{-1mm}
    \item We demonstrate training-free expert pruning across models from 30B to 718B parameters, preserving near-full accuracy at model-dependent sparsity levels of up to 50\%. On DeepSeek-R1, 50\% sparsity halves deployment NPUs and parameters while increasing throughput by 23\%; controlled analyses confirm stability across calibration sources, formats, sample sizes, random seeds, and domain mixtures.
\vspace{-3mm}
\end{itemize}

\section{Related Work}
\vspace{-2mm}

Expert pruning methods target the unique structure of Mixture-of-Experts (MoE) models, where expert utility is often imbalanced across tasks. Prior approaches can be broadly categorized into two paradigms, each with significant limitations. We provide broader context on MoE architectures and general LLM efficiency techniques in Appendix~\ref{app:broader_literature}.

The first paradigm relies on \textit{observation-based statistics}. These methods analyze activation patterns during calibration, using metrics such as activation frequency or gating scores to identify and prune experts~\citep{chen2022task,muzio2024seer}. While straightforward, coarse statistics can fail to distinguish broadly useful generalists from infrequently used but critical specialists. Activation-aware methods provide richer signals: REAP~\citep{lasby2025reap}, for example, combines router gates with expert-output norms for one-shot generative MoE compression. Such signals can improve pruning quality, but require instrumentation and reductions on the expert-output path. PreMoE instead asks how far the router's learned preference alone can go: it profiles scalar router logits before expert dispatch, then filters low-confidence routing noise.

A second paradigm attempts to find optimal expert subsets through computationally intensive \textit{search} procedures. For instance, EEP~\citep{liu2024efficient} employs an evolutionary search, while NAEE~\citep{lu2024not} uses enumeration to evaluate expert combinations based on minimizing reconstruction loss. Although these methods can yield better results than simple heuristics, their computational cost makes them intractable for modern, large-scale MoE models. These search-based techniques are typically demonstrated on models with a small number of experts (e.g., 8 in Mixtral-7B) and would be infeasible for models like DeepSeek-R1 with 256 experts per layer.

PreMoE departs from these paradigms by using refined router logits with confidence filtering to predict expert utility, enabling training-free compilation that scales to large MoE models. 

\section{The PreMoE Framework}
\label{sec:method}

\subsection{Motivation: The Predictability of Expert Activations}
\label{sec:specialization}

\begin{figure*}[htbp]
    \centering
    \includegraphics[width=0.85\linewidth]{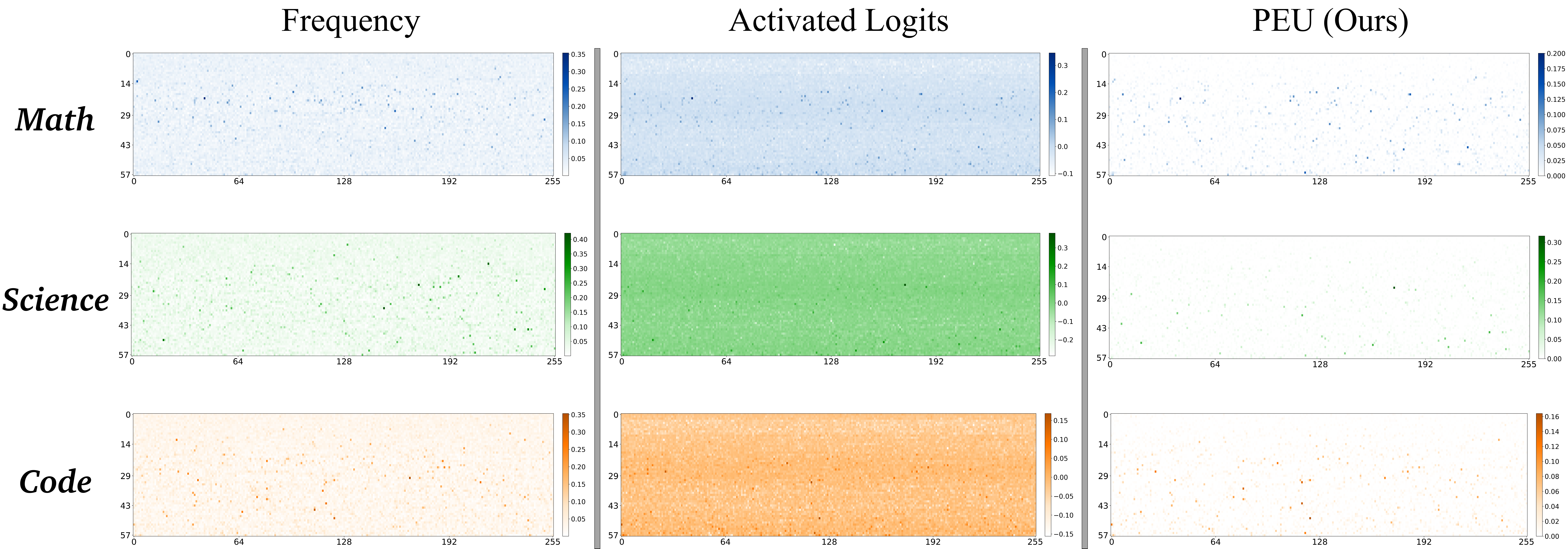} 
    \caption{Comparison of expert utility estimation methods across three domains (\textcolor[RGB]{82,150,214}{Math}, \textcolor[RGB]{0,150,0}{Science}, \textcolor[RGB]{255,120,0}{Code}) on DeepSeek-R1. Each heatmap shows 58 MoE layers (y-axis) by 256 experts (x-axis). \textbf{Left column (Frequency):} Simple activation counting produces diffuse, noisy patterns with minimal domain-specific structure. \textbf{Middle column (Activated Logits):} Aggregating logits only from activated experts yields denser but still broadly distributed signals. \textbf{Right column (Ours):} Our PEU metric, combining high-confidence filtering and logit transformation, reveals sharp, sparse, and highly structured patterns unique to each domain. The contrast demonstrates that expert utility is predictable and domain-specific when measured with a principled signal, motivating our proactive approach.}
    \label{fig:moe_logit_distribution}
\end{figure*}

The foundation of PreMoE rests on a key insight: expert utility in large MoE models is not random, but highly structured and therefore predictable when measured with the right signal.
Figure~\ref{fig:moe_logit_distribution} compares three approaches to estimating expert utility across Math, Science, and Code domains.
In contrast to simple frequency or activated-logits aggregation, our PEU metric reveals sharp, sparse, and highly domain-specific patterns.
These patterns are not only visually distinct across domains but also stable within each domain, indicating that expert utility is predictable when properly measured.
This predictability enables a paradigm shift: from reactive, static deployment to proactive compilation, where we build specialized, efficient model instances tailored to specific deployment scenarios.

\textbf{Notation.} We briefly recap the MoE forward pass to establish notation. For an input token $\mathbf{x}$, a router network produces logits $\mathbf{s}(\mathbf{x}) = \{s_1(\mathbf{x}), \dots, s_{N_r}(\mathbf{x})\}$, where $s_i(\mathbf{x})$ represents the unnormalized preference for the $i$-th routed expert. The top $K$ experts, forming set $\mathcal{K}$, are activated with gating scores $g_i$. The layer output is $F_{\text{MoE}}(\mathbf{x}) = \mathbf{x} + \sum_{i\in\mathcal{K}} g_i\, E_i^{r}(\mathbf{x})$.

\vspace{-4mm}
\subsection{Forecasting Expert Utility}
\label{sec:pep}

To build a proactive system, we need a reliable, low-overhead signal to forecast expert utility. The most natural choice is the router's own logits, as they represent the model's internal, token-by-token preference for each expert.
Crucially, unlike simple activation frequency, logits capture the \textit{strength} of the router's preference. However, raw logits can be noisy. Our goal is to refine this signal to robustly measure the router's \textit{decisive preference}.

To achieve this, we introduce a two-stage process. First, we \textbf{filter for high-confidence activations} using a top-$K_a$ candidate pool and a confidence threshold $r$. Second, for only the logits that pass this filter, we apply a \textbf{logit transformation} $f(s)$ to make them suitable for robust aggregation. This yields a token-level utility score $\tilde{s}_i(\mathbf{x})$ for each expert, formalized as:
\begin{align}
    \mathcal{E}_{K_a} &= \mathrm{TopK}\big(\{s_j(\mathbf{x})\}_{j=1}^{N_r},\, K_a\big) \label{eq:topk}\\
    p_i(\mathbf{x}) &= \frac{\exp\big(s_i(\mathbf{x})\big)}{\sum_{k\in\mathcal{E}_{K_a}} \exp\big(s_k(\mathbf{x})\big)},\quad i\in\mathcal{E}_{K_a} \label{eq:softmax}\\
    \tilde{s}_i(\mathbf{x}) &= \begin{cases}
        f\big(s_i(\mathbf{x})\big), & i\in\mathcal{E}_{K_a}\ \wedge\ p_i(\mathbf{x})\ge r, \\
        0, & \text{otherwise.}
    \end{cases} \label{eq:filtered_score}
\end{align}

\paragraph{Why high-confidence filtering?} Router preference signals are heavy-tailed across tokens; without filtering, aggregations are contaminated by many low-confidence activations that dilute expert utility estimates. Our high-confidence filtering comprises two stages: (1) \textbf{TopK filtering}, which narrows the candidate pool to the top-$K_a$ experts by logit magnitude, and (2) \textbf{adaptive threshold filtering}, which applies a confidence threshold $r$ to retain only decisive activations. Together, these act as a principled denoiser. We set $r$ adaptively for each MoE layer $l$ as the average probability of the top-ranked expert within the top-$K_a$ pool:
\begin{equation}
    r_l = \mathbb{E}_{\mathbf{x} \in \mathcal{X}_T} \left[ \max_{i \in \mathcal{E}_{K_a}^l(\mathbf{x})} p_i^l(\mathbf{x}) \right]
    \label{eq:adaptive_r}
\end{equation}
This layer-wise adaptive rule is computed on the calibration set and makes our method robust across different models and domains with minimal tuning.

\paragraph{Why logit transformation?} Raw router logits are signed, and naive averaging leads to a 0-vs-negative pitfall where unselected experts (scored as 0) can spuriously outrank selected experts with negative logits. A transformation that maps negatives away from the positive range is therefore necessary. Our default choice, $f(s)=\max(s, \mathrm{sigmoid}(s))$, retains large positive evidence while rectifying negatives via $\mathrm{sigmoid}(s)$, avoiding compression of strong signals. In practice, threshold filtering stabilizes utility estimation and largely removes sensitivity to the particular transformation choice, enabling recovery of full-model accuracy at high sparsity.
The final \textbf{Predicted Expert Utility (PEU)} for an expert is the average of these token-level scores over a calibration dataset, $\mathrm{PEU}_i^T = \mathbb{E}_{\mathbf{x} \in \mathcal{X}_T}[\tilde{s}_i(\mathbf{x})]$. This PEU score provides a robust forecast of expert importance for a given domain $T$.
Full derivations are in Appendix~\ref{app:method_details_1}.

\paragraph{From PEU to Computational Patterns.}
By computing PEU scores for all $N_r$ experts across all MoE layers in the model, we obtain a complete ranking of expert importance for domain $T$.
We define this layer-wise collection of PEU scores as the \textbf{computational pattern} for domain $T$:
\begin{equation}
    \text{Pattern}^T = \left\{ \left\{\mathrm{PEU}_i^{T,l}\right\}_{i=1}^{N_r} \right\}_{l=1}^{L}
\end{equation}
where $L$ is the number of MoE layers in the model.
This pattern is a compact, interpretable representation of which experts are critical for a domain, serving as a blueprint for constructing specialized model instances.
Crucially, extracting a pattern requires only one offline forward pass over a small calibration set, with no gradient computation or parameter updates. Five to ten samples already recover most of the final accuracy, while our larger default sets mainly reduce variance in the PEU ranking (Appendix~\ref{app:calibration_efficiency}). The pattern is computed once before deployment and its cost is amortized across inference requests.

\subsection{Compiling Instances from Patterns}
\label{sec:compilation}

Given a computational pattern, we compile specialized MoE instances by populating a lightweight model skeleton \textit{only} with the expert weights identified by the pattern—the full model is never loaded into memory.
The detailed algorithm is provided in Appendix~\ref{app:method_details_2}.

We propose two primary compilation strategies based on computational patterns:

\textbf{1. Compiling Domain-Specific Specialists.} For an application with a narrow focus, we identify the computational pattern for that domain. The top-$M$ experts with the highest PEU scores are selected, where $M$ is the desired budget. These selected experts form the new, pruned set of routed experts $\{E_i^r(\mathbf{x})\}_{i=1}^M$ for the compiled specialist instance.

\textbf{2. Compiling High-Efficiency Generalists.} For general-purpose applications, we create a single, multi-domain model by synthesizing a unified computational pattern. Instead of taking a simple union of top experts, we first average token-level utility within each domain and then combine the domain utilities with weights $w_d$. Our deployment-agnostic default is $w_d=1/D$, so domains contribute equally regardless of calibration-set size; application-specific priors can instead emphasize a target workload. This blended ranking identifies experts useful across the targeted domains, and its top-$M$ experts form a sparse generalist that preserves broad capability under a fixed expert budget. Appendix~\ref{app:mixture_sensitivity} shows that changing the weights produces predictable domain trade-offs.

\begin{table*}[tb!]
    \centering
    \caption{Performance of Domain-Specific Specialists. For each base model we prune routed experts to the highest sparsity that preserves near loss-less accuracy: DeepSeek-R1 at 50\% (keep 128/256 per layer), openPangu-Ultra at 31.25\% (keep 176/256), and Qwen3-30B-A3B at 50\% (keep 64/128). Entries are accuracy (\%); \textbf{Avg} is the macro average over the listed benchmarks; \textbf{$\Delta$} is the average accuracy change compared to the full model. 
    Baselines: Random, Frequency (activation counts), Act-Logits (aggregate logits of activated experts without threshold filtering), All-Logits (aggregate logits of all experts), SEER (L) and SEER (G) from \citet{muzio2024seer}, and EASY-EP.}
    \label{tab:main_specialist_results}
    \resizebox{0.9\textwidth}{!}{%
    \begin{tabular}{l | c c c c c c | c c}
    \toprule
    \textbf{Method} & \textbf{MATH-500} & \textbf{AIME 2024} & \textbf{AIME 2025} & \textbf{CNMO 2024} & \textbf{GPQA} & \textbf{LiveCodeBench} & \textbf{Avg} & \textbf{$\Delta$} \\
    \midrule
    \multicolumn{9}{l}{\textbf{DeepSeek-R1} (50\% sparsity)} \\
    \midrule
    Full & 96.60 & 77.08 & 65.83 & 71.18 & 73.23 & 69.12 & 75.51 & -- \\
    Random & 54.00 & 6.67 & 1.25 & 30.88 & 39.90 & 27.43 & 26.69 & -48.82 \\
    Frequency & 88.80 & 60.83 & 46.19 & 64.58 & 33.33 & 5.88 & 49.93 & -25.58 \\
    All-Logits & 3.60 & 0.00 & 0.00 & 0.35 & 28.79 & 0.00 & 5.46 & -70.05 \\
    SEER (L) & 54.80 & 2.91 & 2.08 & 36.63 & 36.87 & 8.09 & 23.56 & -51.95 \\
    SEER (G) & 53.00 & 3.33 & 1.67 & 36.63 & 36.87 & 14.34 & 24.31 & -51.20 \\
    Act-Logits & 88.20 & 70.00 & 55.00 & 62.62 & 48.48 & 52.94 & 62.87 & -12.64 \\
    EASY-EP & 97.20 & 79.17 & 68.33 & 72.18 & 70.12 & 61.11 & 74.69 & -0.82 \\
    \rowcolor{gray!10} PreMoE & \textbf{97.60} & \textbf{79.58} & \textbf{68.33} & \textbf{75.00} & \textbf{72.22} & \textbf{66.36} & \textbf{76.52} & \textbf{+1.01} \\
    \midrule
    \multicolumn{9}{l}{\textbf{openPangu-Ultra} (31.25\% sparsity)} \\
    \midrule
    Full & 97.40 & 80.83 & 75.42 & 77.43 & 76.77 & 67.65 & 79.25 & -- \\
    Random & 85.00 & 41.25 & 34.17 & 54.51 & 60.10 & 25.74 & 50.13 & -29.12 \\
    Frequency & 96.40 & 76.25 & 65.83 & 75.69 & 52.02 & 51.84 & 69.67 & -9.58 \\
    All-Logits & 12.00 & 0.00 & 0.00 & 2.95 & 16.67 & 0.74 & 5.39 & -73.86 \\
    SEER (L) & 95.20 & 75.41 & 62.08 & 73.61 & 62.63 & 58.82 & 71.29 & -7.96 \\
    SEER (G) & 96.00 & 71.67 & 69.17 & 72.40 & 57.58 & 58.82 & 70.93 & -8.32 \\
    Act-Logits & \textbf{97.20} & 77.50 & \textbf{73.33} & 72.40 & \textbf{79.29} & 61.76 & 76.91 & -2.34 \\
    EASY-EP & 96.82 & 77.20 & 72.17 & 73.43 & 76.74 & 62.36 & 76.45 & -2.80 \\
    \rowcolor{gray!10} PreMoE & 96.80 & \textbf{80.41} & 71.67 & \textbf{79.17} & 75.76 & \textbf{66.91} & \textbf{78.45} & \textbf{-0.80} \\
    \midrule
    \multicolumn{9}{l}{\textbf{Qwen3-30B-A3B} (50\% sparsity)} \\
    \midrule
    Full & 97.20 & 91.25 & 82.92 & 78.65 & 68.69 & 65.44 & 80.69 & -- \\
    Random & 44.80 & 5.00 & 2.92 & 7.99 & 17.68 & 0.00 & 13.07 & -67.62 \\
    Frequency & 90.40 & 60.42 & 46.67 & 54.69 & 47.98 & 25.37 & 54.26 & -32.76 \\
    All-Logits & 1.60 & 0.00 & 0.00 & 0.00 & 0.51 & 0.00 & 0.35 & -80.34 \\
    SEER (L) & 1.60 & 0.00 & 0.00 & 0.00 & 0.00 & 2.21 & 0.64 & -80.05 \\
    SEER (G) & 0.20 & 0.00 & 0.00 & 0.00 & 0.00 & 1.84 & 0.34 & -80.35 \\
    Act-Logits & 1.40 & 0.00 & 0.00 & 0.00 & 3.54 & 0.00 & 0.82 & -79.87 \\
    EASY-EP & 58.62 & 62.46 & 44.92 & 52.65 & 50.48 & 56.44 & 54.26 & -26.43 \\
    \rowcolor{gray!10} PreMoE & \textbf{96.40} & \textbf{88.33} & \textbf{79.58} & \textbf{81.94} & \textbf{68.18} & \textbf{65.07} & \textbf{79.92} & \textbf{-0.77} \\
    \bottomrule
    \end{tabular}%
    }
\end{table*}

\section{Experiments}
\label{sec:exp}

\subsection{Experimental Setup}

\textbf{Models and Datasets.}
Our experiments are conducted on three state-of-the-art MoE models: \textbf{DeepSeek-R1-671B}~\citep{guo2025deepseek} (256 experts per layer), \textbf{openPangu-Ultra-MoE-718B}~\citep{tang2025pangu} (256 experts per layer), and \textbf{Qwen3-30B-A3B} (Qwen3-235B-A22B-Thinking-2507)~\citep{qwen3technicalreport} (128 experts per layer). 
We use a diverse set of open-source training datasets to create calibration sets for three primary domains: 
\textbf{Math} (800 samples from Nemotron-Post-Training-Dataset-v1~\citep{NemotronPostTrainingDatasetV1} ``math'' split), 
\textbf{Science} (200 samples from Nemotron-Post-Training-Dataset-v1~\citep{NemotronPostTrainingDatasetV1} ``stem'' split), and \textbf{Code} (600 samples from OpenCoder~\citep{Huang2024OpenCoderTO}, with 300 samples under opc-sft-stage1 and 300 samples under opc-sft-stage2). 
For the High-Efficiency Generalist, we augment this data with 600 samples from LMSYS-Chat-1M~\citep{zheng2023lmsyschat1m} to broaden its capabilities. 
Samples are drawn randomly without benchmark-specific selection. We collect router statistics on each question together with its full-model reasoning trace under fixed per-model generation settings, covering both prefill and decoding-time routing. Calibration updates no weights and uses no evaluation answers; held-out GSM8K/MBPP validation and prompt-level decontamination checks are reported in Appendix~\ref{app:calibration_validation}. Notably, just 5--10 samples recover most of the final performance, while larger default sets mainly reduce ranking variance (Appendix~\ref{app:calibration_efficiency}).

\textbf{Evaluation Benchmarks.}
To evaluate the performance of the compiled models, we use a challenging set of reasoning benchmarks across our three target domains. 
For \textbf{Math}, we use MATH-500~\citep{hendrycks2020measuring}, AIME 2024, AIME 2025, and CNMO 2024. 
For \textbf{Science}, we use GPQA~\citep{rein2024gpqa}, a graduate-level question-answering benchmark. 
For \textbf{Code}, we use LiveCodeBench~\citep{jain2024livecodebench} (272 problems in the window from 8/1/2024 to 1/1/2025) for evaluating code generation capabilities.

\textbf{Implementation Details}
Pattern collection is performed offline on servers equipped with 64 Ascend 910B2-64GB NPUs. 
For hyperparameters, we set $K_a$ to match the model's default number of activated experts (e.g., 8 for DeepSeek-R1-671B), and we use $f(s) = \max(s, \mathrm{sigmoid}(s))$ as our default logit transformation. 
The confidence threshold $r$ is set adaptively for each domain and for each MoE layer $l$ following the definition in Eq.~\ref{eq:adaptive_r} (see Methodology). This layer-wise adaptive approach makes our method robust across different models and domains with minimal tuning.

\textbf{Baselines.}
We compare PreMoE against a suite of baselines: 
\textbf{Random} selection, 
expert ranking by activation \textbf{Frequency}, 
and several variants of logit collection based on \textbf{all-logits} and \textbf{activated-logits} (denoted `All-Logits' and `Act-Logits' in tables). 
We also compare with local and global \textbf{SEER-MoE}~\citep{muzio2024seer}, \textbf{EASY-EP}~\citep{dong2025domain}, which uses few-shot demonstrations to identify relevant experts, and \textbf{REAP}~\citep{lasby2025reap}, which scores experts by router-weighted activation norms. REAP is evaluated separately under a matched protocol because its experiment reports normalized retention rather than the absolute scores in Table~\ref{tab:main_specialist_results}.

\begin{table*}[tb!]
    \centering
    \caption{Performance of the High-Efficiency Generalist. The PreMoE Generalist is compiled by synthesizing PEU scores at the same sparsity as specialists. The ``Trivial Union'' baseline unions smaller fixed-size per-domain expert sets, yielding lower effective sparsity than PreMoE. Delta ($\Delta$): average accuracy change compared to the full model.}
    \label{tab:main_generalist_results}
    \resizebox{0.9\textwidth}{!}{%
    \begin{tabular}{l | c | c c c c c c | c c}
    \toprule
    \textbf{Setting} & \textbf{Sparsity} & \textbf{MATH-500} & \textbf{AIME 2024} & \textbf{AIME 2025} & \textbf{CNMO 2024} & \textbf{GPQA} & \textbf{LiveCodeBench} & \textbf{Avg} & \textbf{$\Delta$} \\
    \midrule
    \multicolumn{10}{l}{\textbf{DeepSeek-R1}} \\
    \midrule
    Full Model & 0\% & 96.60 & 77.08 & 65.83 & 71.18 & 73.23 & 69.12 & 75.51 & -- \\
    Trivial Union & 39.87\% & 96.60 & 78.75 & 70.42 & 75.17 & 72.22 & 66.18 & 76.56 & +1.05 \\
    \rowcolor{gray!10} PreMoE Generalist & \textbf{50\%} & 96.40 & 78.33 & 70.42 & 75.17 & 70.71 & 65.07 & 76.02 & +0.51 \\
    \midrule
    \multicolumn{10}{l}{\textbf{openPangu-Ultra}} \\
    \midrule
    Full Model & 0\% & 97.40 & 80.83 & 75.42 & 77.43 & 76.77 & 67.65 & 79.25 & -- \\
    Trivial Union & 26.24\% & 97.00 & 80.00 & 70.42 & 80.38 & 74.24 & 63.60 & 77.61 & -1.64 \\
    \rowcolor{gray!10} PreMoE Generalist & \textbf{31.25\%} & 97.40 & 80.00 & 73.34 & 78.82 & 78.79 & 65.81 & 79.03 & -0.22 \\
    \midrule
    \multicolumn{10}{l}{\textbf{Qwen3-30B-A3B}} \\
    \midrule
    Full Model & 0\% & 97.20 & 91.25 & 82.92 & 78.65 & 68.69 & 65.44 & 80.69 & -- \\
    Trivial Union & 33.76\% & 96.40 & 87.92 & 84.17 & 83.33 & 71.72 & 61.76 & 80.88 & +0.19 \\
    \rowcolor{gray!10} PreMoE Generalist & \textbf{50\%} & 96.60 & 86.25 & 79.17 & 79.34 & 68.69 & 54.04 & 77.35 & -3.34 \\
    \bottomrule
    \end{tabular}%
    }
    \vspace{-4mm}
\end{table*}

\subsection{Main Results: Efficacy of PreMoE}

\subsubsection{Compiling Domain-Specific Specialists}
We first evaluate PreMoE's ability to create sparse, high-performance models specialized for a single domain.
For each base model, we first determine a target sparsity using a small sweep with PreMoE, selecting the highest sparsity at which the compiled specialist remains within a 1-point average drop from the full model. We then evaluate all baselines under the same sparsity budget for a fair comparison: 50\% for DeepSeek-R1 (keeping 128 of 256 experts per MoE layer), 31.25\% for openPangu-Ultra-MoE (keeping 176 of 256), and 50\% for Qwen3-30B-A3B (keeping 64 of 128). Table~\ref{tab:main_specialist_results} shows the results.

Across all three models, PreMoE is the only method that consistently matches the full baseline, achieving $\Delta$=+1.01 on DeepSeek-R1, $\Delta$=-0.80 on openPangu-Ultra, and $\Delta$=-0.77 on Qwen3-30B-A3B. The ablation variant Act-Logits---which uses only TopK filtering, without threshold filtering or logit transformation---is substantially less stable, with drops of 12.64 points on DeepSeek-R1, 2.34 points on openPangu-Ultra, and 79.87 points on Qwen3-30B-A3B. This shows that our threshold filtering and logit transformation are not merely incremental refinements, but are crucial for robust expert selection under aggressive sparsity.

Compared with EASY-EP, PreMoE achieves higher average accuracy on all three base models: 76.52 vs.\ 74.69 on DeepSeek-R1, 78.45 vs.\ 76.45 on openPangu-Ultra, and 79.92 vs.\ 54.26 on Qwen3-30B-A3B. The contrast is especially striking on Qwen3-30B-A3B, where EASY-EP suffers a 26.43-point average drop while PreMoE remains near loss-less. In this challenging regime, nearly all alternative baselines fail: on Qwen3-30B-A3B, Random, All-Logits, SEER, and Act-Logits collapse almost entirely, while Frequency still incurs a 32.76-point drop. Overall, these results show that PreMoE can reliably compile domain-specific specialists across a wide range of MoE scales, with the largest advantage appearing precisely where expert sharing makes pruning hardest.

\begin{table}[t!]
    \centering
    \caption{Quality--profiling-cost comparison on DeepSeek-R1 at 50\% sparsity. Accuracy is normalized to the same dense model and averaged over MATH-500, GPQA, and LiveCodeBench. Profiling overhead is measured beyond the shared sparse-MoE calibration forward pass on approximately 600 samples.}
    \label{tab:reap_profile_main}
    \small
    \setlength{\tabcolsep}{4pt}
    \begin{tabular}{lccc}
    \toprule
    \textbf{Method} & \textbf{Profiled signal} & \textbf{Norm. avg.} & \textbf{Overhead} \\
    \midrule
    REAP & router + activation norm & 92.0 & +75\% \\
    PreMoE & scalar router logits & 91.2 & \textbf{+12\%} \\
    \bottomrule
    \end{tabular}
\end{table}

\subsubsection{Compiling High-Efficiency Generalists}
A key advantage of PreMoE is its ability to create a single, sparse model that retains broad, multi-domain capabilities. We compile a generalist model by synthesizing patterns from the Math, Science, and Code domains, rather than simply concatenating per-domain specialist masks. As shown in Table~\ref{tab:main_generalist_results}, the compiled generalist achieves minimal performance loss on the larger models ($\Delta$=+0.51 on DeepSeek-R1; $\Delta$=-0.22 on openPangu-Ultra) at the same sparsity as specialists. On Qwen3-30B-A3B, the generalist shows a larger drop ($\Delta$=-3.34), consistent with the specialist results and suggesting lower expert redundancy at this scale.

We compare to a Trivial Union baseline that unions fixed-size per-domain expert sets. While Trivial Union maintains similar accuracy, it achieves lower effective sparsity, showing that our synthesis approach is more efficient. This indicates that multi-domain compilation is not simply a set-union problem: retaining broad capability requires balancing shared experts with a small number of domain-critical ones. As we show below, this structure also explains why specialists can be highly efficient in-domain while the synthesized generalist offers a better operating point when broader capability retention is required. 

\textbf{Robustness.}
The same specialist and generalist trends hold across alternative calibration sources, random subsets, token formats, and domain mixtures; full controlled results are in Appendix~\ref{app:calibration_robustness}.

\subsubsection{Comparison with Activation-Aware Pruning}
\label{sec:reap_comparison}
REAP combines router gates with expert activation norms. Under the matched 50\%-sparsity protocol in Table~\ref{tab:reap_profile_main}, PreMoE is within 0.8 normalized points of REAP, while its measured profiling overhead is 12\% rather than 75\%. This isolates the benefit and cost of activation-side information: PreMoE uses only scalar router scores, whereas REAP additionally reduces expert outputs. Detailed quality, memory, and cost results are in Appendix~\ref{app:reap_comparison}.

\subsection{Where PreMoE Helps Most: Accuracy-Efficiency Trade-offs}

\begin{wrapfigure}{r}{0.42\linewidth}  
    \centering  
    \includegraphics[width=\linewidth]{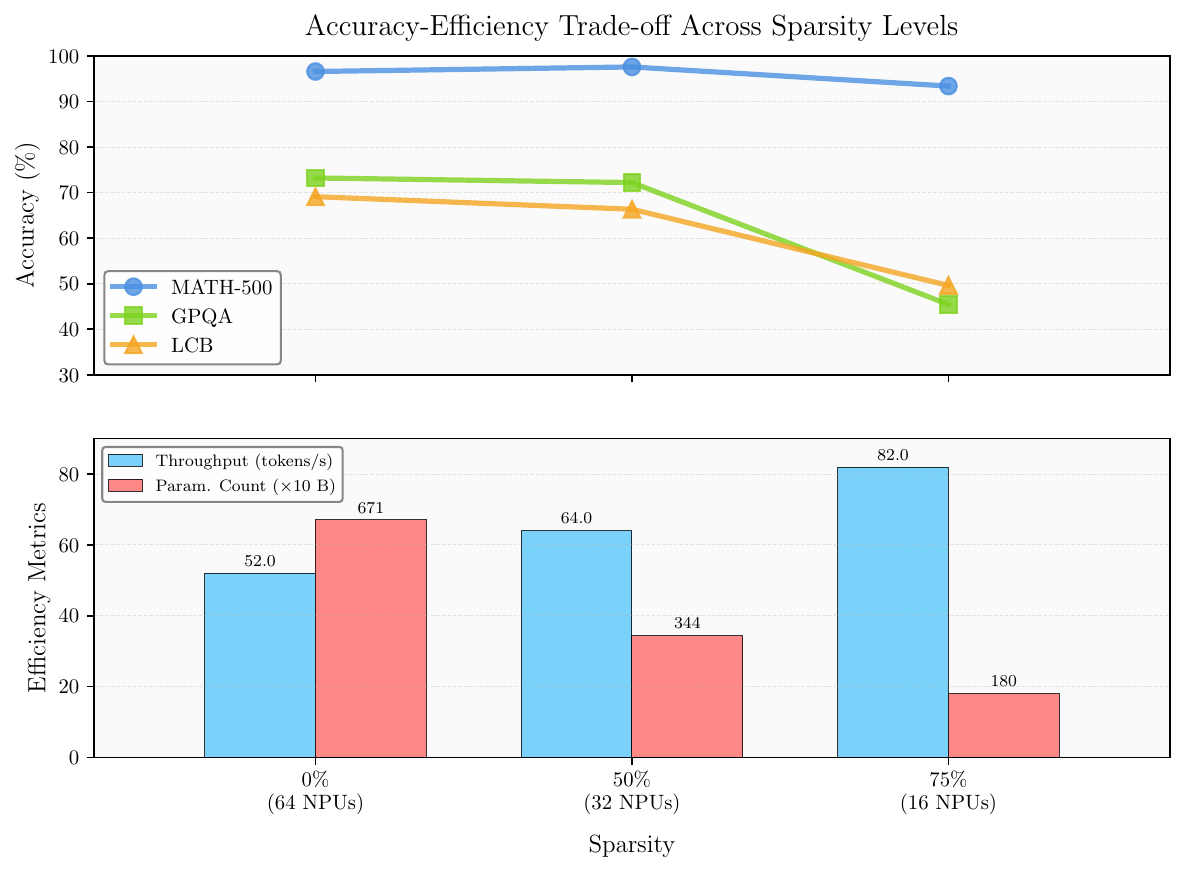}
    \caption{Accuracy-efficiency trade-off: 75\% sparsity preserves MATH accuracy; 50\% sparsity halves infrastructure with 23\% throughput gain.}
    \label{fig:sparsity-tradeoff}
    \vspace{-2mm}  
\end{wrapfigure}

We investigate how performance and deployment efficiency scale with sparsity on DeepSeek-R1. Figure~\ref{fig:sparsity-tradeoff} visualizes the accuracy-efficiency trade-off, revealing domain-dependent robustness and significant infrastructure savings.

\textbf{Domain-Dependent Robustness.}
Mathematical reasoning exhibits remarkable robustness: MATH-500 maintains 96--97\% accuracy even at 62.5\% sparsity. At 75\% sparsity, it retains 93.4\%, while GPQA drops to 45.45\% and LCB to 49.63\%. This suggests that domains differ substantially in expert redundancy: mathematical reasoning appears to rely on a more redundant expert pool, whereas science and code depend on more diverse and specialized capabilities. We therefore do not assume that 50\% pruning is universally lossless: the safe budget is selected per model and workload (50\% for DeepSeek-R1 and Qwen3-30B-A3B, but 31.25\% for openPangu-Ultra). This result is consistent with the overlap analysis above, where only a small subset of top-ranked experts remains strongly domain-specific.

\textbf{Practical Deployment Benefits.}
At 50\% sparsity, we achieve near-lossless accuracy with substantial infrastructure savings: 2× fewer NPUs and parameters, plus 23\% throughput gain. At 75\% sparsity, despite accuracy degradation on some tasks, we gain 4× resource reduction and 58\% throughput increase; a potentially acceptable trade-off for latency-critical applications. Complete accuracy and efficiency metrics at various sparsity levels, including comparisons between domain-specific specialists and multi-domain generalists, are provided in Appendix~\ref{tab:sparsity_specialist}, \ref{tab:sparsity_generalist}, and \ref{tab:efficiency_metrics}.

\subsection{Why PreMoE Works}

\subsubsection{Logit Transformation and Threshold Filtering}

\begin{figure}[H]
    \centering
    \begin{subfigure}[t]{0.34\linewidth}
        \centering
        \includegraphics[width=\linewidth]{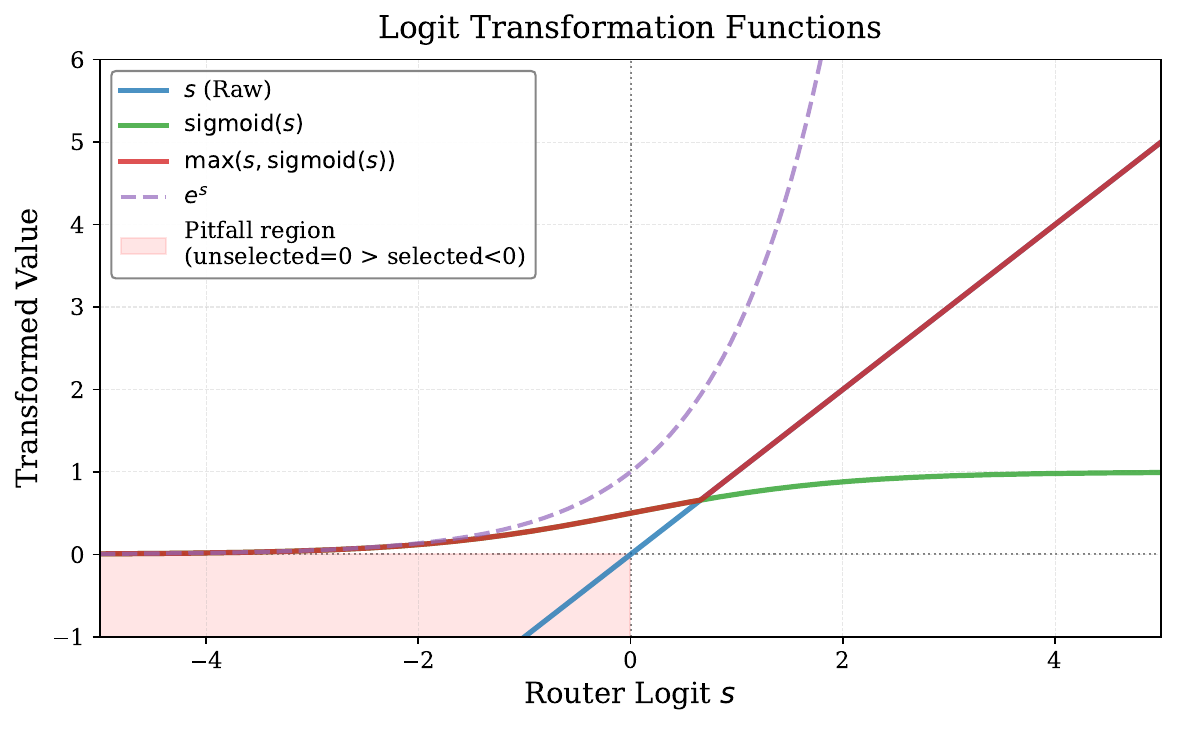}
        \caption{Transformation functions}
        \label{fig:logit-trans-a}
    \end{subfigure}\hfill
    \begin{subfigure}[t]{0.65\linewidth}
        \centering
        \includegraphics[width=\linewidth]{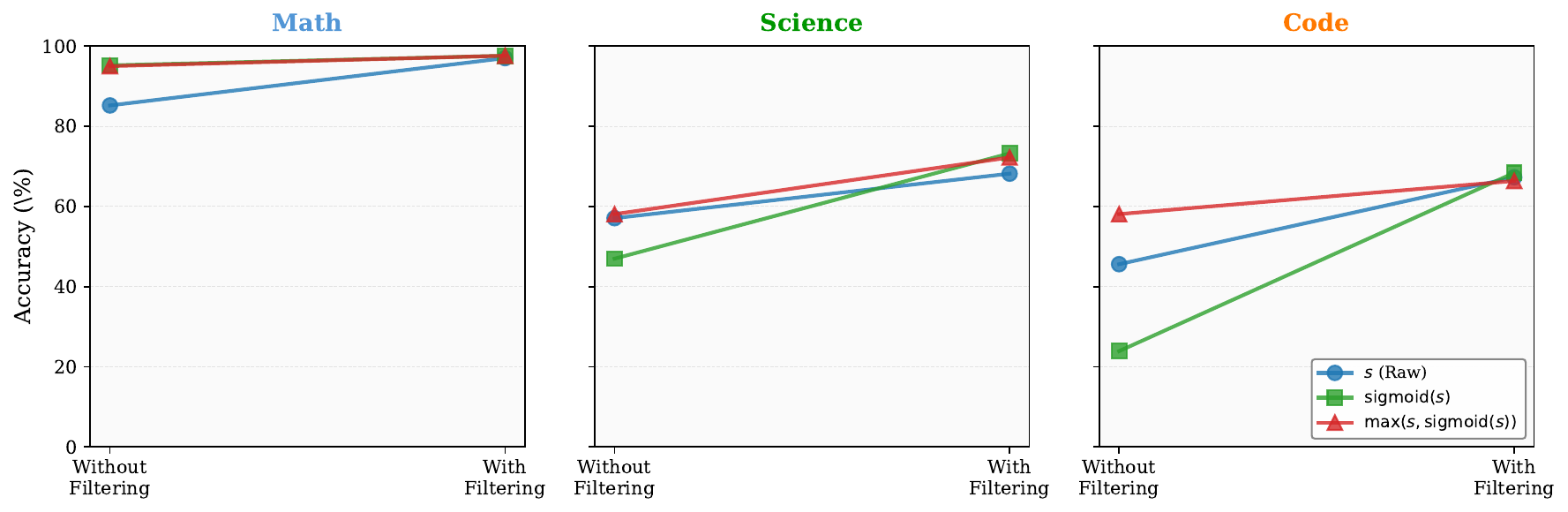}
        \caption{Convergence with filtering across domains}
        \label{fig:logit-trans-b}
    \end{subfigure}
    \caption{Analysis of logit transformation and threshold filtering. (a) Transformation curves illustrating the 0-vs-negative issue; the shaded region marks cases where unselected experts (score 0) outrank selected experts with negative logits. (b) Convergence across \textcolor[RGB]{82,150,214}{Math}, \textcolor[RGB]{0,150,0}{Science}, and \textcolor[RGB]{255,120,0}{Code}. Without threshold filtering, performance depends strongly on the transformation; with filtering, all methods converge, showing that thresholding is the main factor in stabilizing utility estimation.}
    \label{fig:logit-transformation}
    \vspace{-4mm}
\end{figure}

All experiments in this subsection use TopK filtering as the baseline, on top of which we study threshold filtering and logit transformation. Raw router logits \(s\) can be negative or positive. When utilities are aggregated across tokens, unselected experts receive a default score of 0, which can incorrectly exceed the scores of selected experts with negative logits. Averaging raw logits is therefore problematic. To address this, we transform logits before aggregation. A natural choice is \(\sigma(s)\), which maps logits to \((0,1)\) and avoids the sign issue, but it also compresses large positive values. Our default rectifier, $f(s)=\max(s,\sigma(s))$ uses \(\sigma(s)\) for negative logits while preserving positive logits unchanged. We also evaluate an exponential variant \(f_{\exp}(s)=e^s\); additional results are deferred to Appendix~\ref{app:exp_amplifier}.

Figure~\ref{fig:logit-transformation} shows two main findings. First, without threshold filtering, the choice of logit transformation matters: \(f(s)=\max(s,\sigma(s))\) is the most consistent across domains, while \(\sigma(s)\) performs well on Math but degrades on Science and Code. Second, once threshold filtering is introduced, all transformations achieve similarly strong performance. This indicates that logit transformation mainly corrects the negative-logit scoring issue, whereas threshold filtering is the primary factor in stabilizing utility estimation. We therefore use \(\max(s,\sigma(s))\) as the default.

We further ablate the thresholding strategy. On DeepSeek-R1 at 50\% sparsity, the adaptive threshold achieves \(98.0/64.71/70.71\) on MATH-500/LCB/GPQA, outperforming fixed thresholds of 0.15 \((94.2/64.34/68.69)\) and 0.3 \((96.2/65.07/68.18)\). A full component ablation is provided in Appendix~\ref{app:component_ablation}.



\subsubsection{Diverse and Critical Expert Roles}
Analyzing the PEU rankings on the Code domain reveals distinct expert roles, which in turn clarify why frequency-based pruning fails under aggressive sparsity.

\textbf{Frequently-Activated Generalists (Low Utility).} 
Some experts are considered by the router thousands of times but are rarely the decisive top choice. Table~\ref{tab:case1_experts} shows five such examples from the Code domain. Frequency-based methods rank these in the top-63, while PEU correctly demotes them to positions 252 to 256. These are ``frequent but weak'' generalists that occupy model capacity without providing critical utility for code generation tasks.

\textbf{Infrequent Specialists (High Contextual Utility).} 
Conversely, some experts are rarely activated but are almost always the top choice when they are. Table~\ref{tab:case2_experts} shows five such examples. Frequency-based methods rank these at the tail, effectively discarding them, while PEU promotes them to positions 51 to 148. These are rare-but-critical specialists whose removal would eliminate essential capabilities for specific code constructs or programming patterns that occur infrequently in the calibration set but are vital when they appear.

\begin{table*}[tb!]
    \centering
    \caption{Expert role diversity. \textbf{Left:} Frequently activated generalists with low utility: often selected, but rarely the top choice, so frequency overestimates them while PEU demotes them. \textbf{Right:} Infrequent specialists with high utility: rarely selected, but often the top choice when activated, so frequency overlooks them while PEU promotes them.}
    \label{tab:expert_roles}
    \begin{minipage}[t]{0.45\textwidth}
        \centering
        \subcaption{Frequently-Activated Generalists (Low Utility)}
        \label{tab:case1_experts}
        \vspace{0.2em}
        \resizebox{\textwidth}{!}{%
        \begin{tabular}{l | r r r | r r}
        \toprule
        \textbf{Expert} & \textbf{\#Top-$K_a$} & \textbf{\#Rank-1} & \textbf{R-1 Ratio} & \textbf{Freq.} & \textbf{PEU} \\
        \midrule
        (0, 161) & 31182 & 7 & 0.02\% & 140 & 256 \\
        (57, 205) & 42802 & 0 & 0.00\% & 63 & 252 \\
        (2, 54) & 48168 & 3 & 0.01\% & 22 & 256 \\
        (6, 158) & 61977 & 92 & 0.15\% & 10 & 254 \\
        (57, 24) & 47556 & 0 & 0.00\% & 53 & 253 \\
        \bottomrule
        \end{tabular}%
        }
    \end{minipage}%
    \hfill
    \begin{minipage}[t]{0.45\textwidth}
        \centering
        \subcaption{Infrequent Specialists (High Contextual Utility)}
        \label{tab:case2_experts}
        \vspace{0.2em}
        \resizebox{\textwidth}{!}{%
        \begin{tabular}{l | r r r | r r}
        \toprule
        \textbf{Expert} & \textbf{\#Top-$K_a$} & \textbf{\#Rank-1} & \textbf{R-1 Ratio} & \textbf{Freq.} & \textbf{PEU} \\
        \midrule
        (3, 243) & 8382 & 4979 & 59.4\% & 256 & 51 \\
        (36, 9) & 2518 & 1482 & 58.9\% & 255 & 77 \\
        (45, 220) & 640 & 368 & 57.5\% & 256 & 137 \\
        (50, 223) & 655 & 352 & 53.7\% & 256 & 148 \\
        (57, 95) & 8254 & 4370 & 53.0\% & 227 & 58 \\
        \bottomrule
        \end{tabular}%
        }
    \end{minipage}
    \vspace{-4mm}
\end{table*}

These contrasting cases explain why PreMoE maintains performance at high sparsity while frequency-based methods fail: PEU preserves mission-critical specialists while pruning low-utility generalists, whereas frequency does the opposite. This distinction is critical: when we prune to 50\% sparsity on DeepSeek-R1 (keeping 128/256 experts), PEU retains the high-utility specialists while Frequency-based methods lose them, explaining the 60.48 percentage point accuracy gap on Code (66.36\% vs. 5.88\% in Table~\ref{tab:main_specialist_results}).


\Needspace{4\baselineskip}
\textbf{Cross-Domain Expert Overlap.} 

\begin{figure}[tb!]  
    \centering
    \begin{minipage}{0.38\linewidth}
        \centering
        \includegraphics[width=\linewidth]{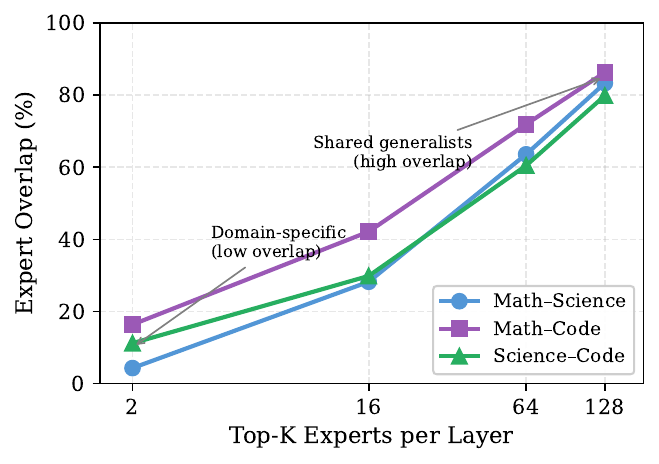}
        \caption{Top-2 specialists exhibit <17\% overlap, while top-128 sets share 80-86\% experts, revealing strong specialization.}
        \label{fig:expert-overlap}
    \end{minipage}
    \hfill  
    \begin{minipage}{0.54\linewidth}
        \centering
        \caption{Cross-domain performance of DeepSeek-R1 specialists at 50\% sparsity. \textbf{Bold} = in-domain; specialists excel in-domain but degrade sharply out-of-domain.}
        \label{tab:ood_specialists}
        \resizebox{\linewidth}{!}{
        \begin{tabular}{l|cccc}
        \toprule
        \textbf{Model} & \textbf{MATH-500} & \textbf{LCB} & \textbf{GPQA} & \textbf{MMLU-Pro} \\
        \midrule
        Full & 96.6 & 69.12 & 73.23 & 82.30 \\
        \midrule
        Math Specialist & \textbf{97.6} & 58.46 & 59.09 & 62.82 \\
        Code Specialist & 87.8 & \textbf{66.36} & 40.91 & 56.71 \\
        \bottomrule
        \end{tabular}%
        }
    \end{minipage}
    \vspace{-2mm}  
\end{figure}

To quantify expert domain-specificity across utility levels, we measured overlap between domain specialists' top-ranked experts on DeepSeek-R1. Figure~\ref{fig:expert-overlap} shows overlap declines sharply with expert criticality: at top-128 (50\% sparsity), domains share 80--86\% of experts; at top-16, this falls to 28--42\%; at top-2, only 4--16\% overlap remains. This confirms top experts are highly domain-specific, while lower-ranked experts act as shared generalists—aligning with our prior expert-role analysis (highest-utility experts exhibit strongest domain specificity). This layered expert-sharing structure explains why aggressive pruning is challenging: broad capability relies on shared backbone retention, while peak in-domain performance depends on a small set of domain-specific experts. This pattern further clarifies why domain specialists outperform on target tasks and why our generalist synthesis requires careful blending of domain patterns (rather than simple union).

\subsubsection{Out-of-Distribution Generalization}

A natural question is how compiled models generalize beyond calibration. Table~\ref{tab:ood_specialists} clarifies the intended use of specialists: they maximize targeted efficiency, but can degrade sharply elsewhere; for example, the Science Specialist drops to 8.46\% on LCB versus 69.12\% for the full model. This trade-off motivates the generalist, which reaches 71.36\% on the broad MMLU-Pro benchmark and improves to 77.58\% when calibration covers additional domains. Thus, specialists provide peak in-domain efficiency while generalists offer a better operating point for broad capability retention. A static profile library is another option for known domains, although automatic shift detection and profile selection remain future work. Full breakdowns and the oracle upper bound are in Appendices~\ref{app:mmlu_pro} and~\ref{app:profile_library}.

\section{Conclusion}

This work introduced PreMoE, a training-free framework that proactively compiles MoE instances for target deployment scenarios. Its Predicted Expert Utility metric extracts domain-specific computational patterns from native router logits through high-confidence filtering and logit transformation. These patterns compile both efficient specialists and broad generalists without retraining. Across models from 30B to 718B parameters, PreMoE preserves near-full accuracy at model-appropriate sparsity levels of up to 50\%; on DeepSeek-R1, 50\% sparsity halves deployment NPUs and parameters while increasing throughput by 23\%. Controlled calibration studies confirm that the learned patterns are stable across sources and random subsets, while also making the intended deployment boundary explicit. PreMoE therefore turns the routing structure already learned by large MoE models into a practical tool for deployment-specific efficiency.

%

\newpage
\bibliography{custom,ref/main}

@article{lasby2025reap,
  title={{REAP} the Experts: Why Pruning Prevails for One-Shot {MoE} Compression},
  author={Lasby, Mike and Lazarevich, Ivan and Sinnadurai, Nish and Lie, Sean and Ioannou, Yani and Thangarasa, Vithursan},
  journal={arXiv preprint arXiv:2510.13999},
  year={2025},
  url={https://arxiv.org/abs/2510.13999}
}

@inproceedings{roller2021hash,
  title={Hash Layers for Large Sparse Models},
  author={Roller, Stephen and Sukhbaatar, Sainbayar and Szlam, Arthur and Weston, Jason},
  booktitle={Advances in Neural Information Processing Systems},
  volume={34},
  pages={17555--17566},
  year={2021}
}

@article{lowell2024alignment,
  title={Training on the Edge of Stability Is Caused by Layerwise Jacobian Alignment},
  author={Lowell, Mark and Kastner, Catharine},
  journal={arXiv preprint arXiv:2406.00127},
  year={2024},
  url={https://arxiv.org/abs/2406.00127}
}

@misc{qwen3technicalreport,
      title={Qwen3 Technical Report}, 
      author={Qwen Team},
      year={2025},
      eprint={2505.09388},
      archivePrefix={arXiv},
      primaryClass={cs.CL},
      url={https://arxiv.org/abs/2505.09388}, 
}

@article{jain2024livecodebench,
  title={Livecodebench: Holistic and contamination free evaluation of large language models for code},
  author={Jain, Naman and Han, King and Gu, Alex and Li, Wen-Ding and Yan, Fanjia and Zhang, Tianjun and Wang, Sida and Solar-Lezama, Armando and Sen, Koushik and Stoica, Ion},
  journal={arXiv preprint arXiv:2403.07974},
  year={2024}
}

@misc{zheng2023lmsyschat1m,
      title={LMSYS-Chat-1M: A Large-Scale Real-World LLM Conversation Dataset}, 
      author={Lianmin Zheng and Wei-Lin Chiang and Ying Sheng and Tianle Li and Siyuan Zhuang and Zhanghao Wu and Yonghao Zhuang and Zhuohan Li and Zi Lin and Eric. P Xing and Joseph E. Gonzalez and Ion Stoica and Hao Zhang},
      year={2023},
      eprint={2309.11998},
      archivePrefix={arXiv},
      primaryClass={cs.CL}
}

@misc{Huang2024OpenCoderTO,
  title = {OpenCoder: The Open Cookbook for Top-Tier Code Large Language Models},
  author = {Siming Huang and Tianhao Cheng and Jason Klein Liu and Jiaran Hao and Liuyihan Song and Yang Xu and J. Yang and J. H. Liu and Chenchen Zhang and Linzheng Chai and Ruifeng Yuan and Zhaoxiang Zhang and Jie Fu and Qian Liu and Ge Zhang and Zili Wang and Yuan Qi and Yinghui Xu and Wei Chu},
  year = {2024},
  url = {https://arxiv.org/pdf/2411.04905}
}

@misc{NemotronPostTrainingDatasetV1,
      author = {Nathawani, Dhruv and Gitman, Igor and Majumdar, Somshubra and Bakhturina, Evelina and Mahabaleshwarkar, Ameya Sunil and Zhang, Jian and Polak Scowcroft, Jane},
      title = {{Nemotron-Post-Training-Dataset-v1}},
      version = {1.0},
      publisher = {{NVIDIA}},
      year = {2025}, month = {July},
      url = {https://huggingface.co/datasets/nvidia/Nemotron-Post-Training-Dataset-v1}
}

@article{chen2022task,
  title={Task-specific expert pruning for sparse mixture-of-experts},
  author={Chen, Tianyu and Huang, Shaohan and Xie, Yuan and Jiao, Binxing and Jiang, Daxin and Zhou, Haoyi and Li, Jianxin and Wei, Furu},
  journal={arXiv preprint arXiv:2206.00277},
  year={2022}
}

@article{liu2024efficient,
  title={Efficient expert pruning for sparse mixture-of-experts language models: Enhancing performance and reducing inference costs},
  author={Liu, Enshu and Zhu, Junyi and Lin, Zinan and Ning, Xuefei and Blaschko, Matthew B and Yan, Shengen and Dai, Guohao and Yang, Huazhong and Wang, Yu},
  journal={arXiv preprint arXiv:2407.00945},
  year={2024}
}

@article{muzio2024seer,
  title={Seer-moe: Sparse expert efficiency through regularization for mixture-of-experts},
  author={Muzio, Alexandre and Sun, Alex and He, Churan},
  journal={arXiv preprint arXiv:2404.05089},
  year={2024}
}

@article{lu2024not,
  title={Not all experts are equal: Efficient expert pruning and skipping for mixture-of-experts large language models},
  author={Lu, Xudong and Liu, Qi and Xu, Yuhui and Zhou, Aojun and Huang, Siyuan and Zhang, Bo and Yan, Junchi and Li, Hongsheng},
  journal={arXiv preprint arXiv:2402.14800},
  year={2024}
}

@article{shazeer2017outrageously,
  title={Outrageously large neural networks: The sparsely-gated mixture-of-experts layer},
  author={Shazeer, Noam and Mirhoseini, Azalia and Maziarz, Krzysztof and Davis, Andy and Le, Quoc and Hinton, Geoffrey and Dean, Jeff},
  journal={arXiv preprint arXiv:1701.06538},
  year={2017}
}

@article{tang2025pangu,
  title={Pangu Ultra MoE: How to Train Your Big MoE on Ascend NPUs},
  author={Tang, Yehui and Yin, Yichun and Wang, Yaoyuan and Zhou, Hang and Pan, Yu and Guo, Wei and Zhang, Ziyang and Rang, Miao and Liu, Fangcheng and Zhang, Naifu and others},
  journal={arXiv preprint arXiv:2505.04519},
  year={2025}
}

@article{pei2024fusegpt,
  title={FuseGPT: Learnable Layers Fusion of Generative Pre-trained Transformers},
  author={Pei, Zehua and Zhen, Hui-Ling and Yu, Xianzhi and Pan, Sinno Jialin and Yuan, Mingxuan and Yu, Bei},
  journal={arXiv preprint arXiv:2411.14507},
  year={2024}
}

@inproceedings{rein2024gpqa,
  title={Gpqa: A graduate-level google-proof q\&a benchmark},
  author={Rein, David and Hou, Betty Li and Stickland, Asa Cooper and Petty, Jackson and Pang, Richard Yuanzhe and Dirani, Julien and Michael, Julian and Bowman, Samuel R},
  booktitle={First Conference on Language Modeling},
  year={2024}
}

@article{zhou2022mixture,
  title={Mixture-of-experts with expert choice routing},
  author={Zhou, Yanqi and Lei, Tao and Liu, Hanxiao and Du, Nan and Huang, Yanping and Zhao, Vincent and Dai, Andrew M and Le, Quoc V and Laudon, James and others},
  journal={Advances in Neural Information Processing Systems},
  volume={35},
  pages={7103--7114},
  year={2022}
}

@article{hinton2015distilling,
  title={Distilling the knowledge in a neural network},
  author={Hinton, Geoffrey and Vinyals, Oriol and Dean, Jeff},
  journal={arXiv preprint arXiv:1503.02531},
  year={2015}
}

@article{hoefler2021sparsity,
  title={Sparsity in deep learning: Pruning and growth for efficient inference and training in neural networks},
  author={Hoefler, Torsten and Alistarh, Dan and Ben-Nun, Tal and Dryden, Nikoli and Peste, Alexandra},
  journal={Journal of Machine Learning Research},
  volume={22},
  number={241},
  pages={1--124},
  year={2021}
}

@article{frankle2018lottery,
  title={The lottery ticket hypothesis: Finding sparse, trainable neural networks},
  author={Frankle, Jonathan and Carbin, Michael},
  journal={arXiv preprint arXiv:1803.03635},
  year={2018}
}

@article{frantar2022gptq,
  title={Gptq: Accurate post-training quantization for generative pre-trained transformers},
  author={Frantar, Elias and Ashkboos, Saleh and Hoefler, Torsten and Alistarh, Dan},
  journal={arXiv preprint arXiv:2210.17323},
  year={2022}
}

@article{dettmers2022gpt3,
  title={Gpt3. int8 (): 8-bit matrix multiplication for transformers at scale},
  author={Dettmers, Tim and Lewis, Mike and Belkada, Younes and Zettlemoyer, Luke},
  journal={Advances in neural information processing systems},
  volume={35},
  pages={30318--30332},
  year={2022}
}

@article{guo2025deepseek,
  title={Deepseek-r1: Incentivizing reasoning capability in llms via reinforcement learning},
  author={Guo, Daya and Yang, Dejian and Zhang, Haowei and Song, Junxiao and Zhang, Ruoyu and Xu, Runxin and Zhu, Qihao and Ma, Shirong and Wang, Peiyi and Bi, Xiao and others},
  journal={arXiv preprint arXiv:2501.12948},
  year={2025}
}

@article{lepikhin2020gshard,
  title={Gshard: Scaling giant models with conditional computation and automatic sharding},
  author={Lepikhin, Dmitry and Lee, HyoukJoong and Xu, Yuanzhong and Chen, Dehao and Firat, Orhan and Huang, Yanping and Krikun, Maxim and Shazeer, Noam and Chen, Zhifeng},
  journal={arXiv preprint arXiv:2006.16668},
  year={2020}
}

@article{dai2024deepseekmoe,
  title={Deepseekmoe: Towards ultimate expert specialization in mixture-of-experts language models},
  author={Dai, Damai and Deng, Chengqi and Zhao, Chenggang and Xu, RX and Gao, Huazuo and Chen, Deli and Li, Jiashi and Zeng, Wangding and Yu, Xingkai and Wu, Y and others},
  journal={arXiv preprint arXiv:2401.06066},
  year={2024}
}

@article{fedus2022switch,
  title={Switch transformers: Scaling to trillion parameter models with simple and efficient sparsity},
  author={Fedus, William and Zoph, Barret and Shazeer, Noam},
  journal={Journal of Machine Learning Research},
  volume={23},
  number={120},
  pages={1--39},
  year={2022}
}

@article{hendrycks2020measuring,
  title={Measuring massive multitask language understanding},
  author={Hendrycks, Dan and Burns, Collin and Basart, Steven and Zou, Andy and Mazeika, Mantas and Song, Dawn and Steinhardt, Jacob},
  journal={arXiv preprint arXiv:2009.03300},
  year={2020}
}

@article{pei2025cmoe,
  title={CMoE: Fast Carving of Mixture-of-Experts for Efficient LLM Inference},
  author={Pei, Zehua and Zou, Lancheng and Zhen, Hui-Ling and Yu, Xianzhi and Liu, Wulong and Pan, Sinno Jialin and Yuan, Mingxuan and Yu, Bei},
  journal={arXiv preprint arXiv:2502.04416},
  year={2025}
}

@article{jiang2024mixtral,
  title={Mixtral of experts},
  author={Jiang, Albert Q and Sablayrolles, Alexandre and Roux, Antoine and Mensch, Arthur and Savary, Blanche and Bamford, Chris and Chaplot, Devendra Singh and Casas, Diego de las and Hanna, Emma Bou and Bressand, Florian and others},
  journal={arXiv preprint arXiv:2401.04088},
  year={2024}
}

@article{dong2025domain,
  title={Domain-specific pruning of large mixture-of-experts models with few-shot demonstrations},
  author={Dong, Zican and Peng, Han and Liu, Peiyu and Zhao, Wayne Xin and Wu, Dong and Xiao, Feng and Wang, Zhifeng},
  journal={arXiv preprint arXiv:2504.06792},
  year={2025}
}
\bibliographystyle{colm2026_conference}

\clearpage
\appendix
\counterwithin{figure}{section}
\counterwithin{table}{section}

\section{Detailed Methodology}
\label{app:method_details}

This section provides the detailed mathematical formulation for the PreMoE framework, as referenced in Section~\ref{sec:method}.

\subsection{Detailed Formulation of Predicted Expert Utility (PEU)}
\label{app:method_details_1}

As described in the main text, the calculation of the Predicted Expert Utility (PEU) is a two-stage process designed to refine the raw router logit signal into a robust measure of an expert's utility.

\textbf{Stage 1: Filtering for High-Confidence Activations.}
The first stage filters out low-confidence or irrelevant expert considerations on a per-token basis. This process is governed by two hyperparameters: $K_a$, the size of an initial candidate pool, and $r$, a confidence threshold. For each token $\mathbf{x}$, we first identify the set of $K_a$ experts with the top raw router logits $s_i(\mathbf{x})$, denoted as $\mathcal{E}_{K_a}(\mathbf{x})$. Next, we compute locally normalized probabilities $p_i(\mathbf{x})$ for this candidate set by applying a softmax function only over their logits. A confidence threshold $r$ is then applied to these local probabilities. An expert is only considered for the next stage if it is in the candidate pool $\mathcal{E}_{K_a}$ and its local probability $p_i(\mathbf{x})$ exceeds the threshold $r$.

The adaptive threshold $r_l$ for each MoE layer $l$ is calculated as the average probability of the top-ranked expert within the top-$K_a$ pool across the calibration dataset:
\begin{equation}
    r_l = \mathbb{E}_{\mathbf{x} \in \mathcal{X}_T} \left[ \max_{i \in \mathcal{E}_{K_a}^l(\mathbf{x})} p_i^l(\mathbf{x}) \right]
\end{equation}
This layer-wise adaptive rule makes our method robust across different models and domains with minimal tuning.

\textbf{Stage 2: Applying the Logit Transformation.}
For the high-confidence expert logits that pass the filtering stage, we then apply the logit transformation $f(s)$. Our default choice is $f(s)=\max(s, \mathrm{sigmoid}(s))$, which retains large positive evidence while rectifying negatives via $\mathrm{sigmoid}(s)$, avoiding compression of strong signals.

The final token-level score, $\tilde{s}_i(\mathbf{x})$, combines these two stages. The full calculation, along with the final PEU averaging, is as follows:
\begin{align}
    \mathcal{E}_{K_a} &= \mathrm{TopK}\big(\{s_j(\mathbf{x})\}_{j=1}^{N_r},\, K_a\big) \\
    p_i(\mathbf{x}) &= \frac{\exp\big(s_i(\mathbf{x})\big)}{\sum_{k\in\mathcal{E}_{K_a}} \exp\big(s_k(\mathbf{x})\big)},\quad i\in\mathcal{E}_{K_a} \\
    \tilde{s}_i(\mathbf{x}) &= \begin{cases}
        f\big(s_i(\mathbf{x})\big), & i\in\mathcal{E}_{K_a}\ \wedge\ p_i(\mathbf{x})\ge r, \\
        0, & \text{otherwise.}
    \end{cases} \\
    \mathrm{PEU}_i^T &= \frac{1}{|\mathcal{X}_T|} \sum_{\mathbf{x} \in \mathcal{X}_T} \tilde{s}_{i}(\mathbf{x}).
\end{align}
This final PEU score represents the average, high-confidence, transformed logit for expert $i$ on the calibration dataset, and is used to rank and select experts for the final computational pattern.

\subsection{Detailed Algorithm for Compiling Specialized Instances}
\label{app:method_details_2}

As described in the main text, the identified computational patterns serve as blueprints for compiling specialized MoE instances. The key to this process is its memory efficiency: the full, dense model (which can be hundreds of gigabytes) is never loaded into memory. Instead, a lightweight ``skeleton'' of the model architecture is first instantiated. Then, only the weights for the selected experts (identified by the PEU-based pattern) are loaded from storage and placed into the appropriate slots in the model. This creates a sparse, powerful, and ready-to-use model instance without ever incurring the memory cost of the full model. This subsection provides the detailed algorithm for the two primary compilation strategies based on this principle.

\textbf{1. Compiling a Domain-Specific Specialist.}
This is the most straightforward application of a computational pattern.
\begin{enumerate}
    \item \textbf{Pattern Identification:} For a target domain $T$, calculate the PEU scores $\{\mathrm{PEU}_i^T\}_{i=1}^{N_r}$ for all experts in each MoE layer across a calibration dataset $\mathcal{X}_T$. This forms the computational pattern for the domain.
    \item \textbf{Expert Selection:} For a given expert budget $M$, select the set of $M$ experts with the highest PEU scores for each layer. This pruned set of experts becomes the new set of routed experts $\{E_i^r(\mathbf{x})\}_{i=1}^M$ for the compiled instance.
    \item \textbf{Instance Compilation:} A new, sparse model instance is created containing only the selected routed experts. The router weights are also pruned to remove parameters corresponding to the discarded experts.
\end{enumerate}

\textbf{2. Compiling a High-Efficiency Generalist.}
This strategy creates a single, sparse model that retains capability across multiple domains by creating a synthesized, multi-domain computational pattern.
\begin{enumerate}
    \item \textbf{Synthesize Token-Level Scores:} For a set of $D$ target domains $\{T_1, \ldots, T_D\}$, collect the token-level utility scores, $\{\tilde{s}_i(\mathbf{x})\}$, from their calibration datasets $\{\mathcal{X}_{T_1}, \ldots, \mathcal{X}_{T_D}\}$.
    \item \textbf{Calculate Multi-Domain PEU:} We first average within each domain and then combine domains with deployment weights $w_d$:
    \begin{equation}
        \mathrm{PEU}_i^{\text{multi}} = \sum_{d=1}^D w_d
        \left(\frac{1}{|\mathcal{X}_{T_d}|}\sum_{\mathbf{x}\in\mathcal{X}_{T_d}}\tilde{s}_i(\mathbf{x})\right),
        \qquad \sum_{d=1}^D w_d=1.
    \end{equation}
    Our deployment-agnostic default is $w_d=1/D$, preventing a larger calibration set from implicitly dominating the pattern. Other weights encode application-specific workload priorities.
    \item \textbf{Expert Selection:} For a given total expert budget $M$, select the set of $M$ experts with the highest multi-domain PEU scores. This becomes the new set of routed experts $\{E_i^r(\mathbf{x})\}_{i=1}^M$.
    \item \textbf{Instance Compilation:} A new model instance is created containing only the final selected set of routed experts.
\end{enumerate}
This compilation process is performed once at deployment time, creating a static, efficient model instance that is proactively specialized for its intended application, whether that be single-domain or multi-domain.

\subsection{Generation Configuration}
\label{app:generation_config}

For reproducibility, we provide the generation configurations used for each model during both calibration (pattern collection) and evaluation:

\begin{table}[H]
    \centering
    \caption{Generation configurations for each model.}
    \label{tab:generation_config}
    \small
    \setlength{\tabcolsep}{8pt}
    \begin{tabular}{l|ccc}
    \toprule
    \textbf{Model} & \textbf{Temperature} & \textbf{Top-p} & \textbf{Top-K} \\
    \midrule
    DeepSeek-R1 & 0.6 & 0.95 & -- \\
    openPangu-Ultra & 0.7 & 1.0 & -- \\
    Qwen3-30B-A3B & 0.6 & 0.95 & 20 \\
    \bottomrule
    \end{tabular}
\end{table}

For all models, we use a maximum context length (input + output) of 32,768 tokens during evaluation. The exception is Qwen3-30B-A3B on AIME 2024, AIME 2025, and CNMO 2024, where we use 38,912 tokens following the official evaluation setting from the Qwen3 technical report.

\section{Additional Attempt: Exponential Amplifier}
\label{app:exp_amplifier}
Complementing the logit transformation choices described in the main text, we also investigated an exponential amplifier as an additional attempt:
\begin{equation}
    f_{\exp}(s) = e^{s}.
\end{equation}
The motivation is to aggressively enlarge gaps among activated experts, which can be beneficial when many low-confidence activations introduce noise and raw logits contain both negative and positive values.

\textbf{Protocol.} We evaluated $f_{\exp}$ in two settings: (i) applied directly to the top-$K_a$ activated experts without threshold filtering, and (ii) applied after the high-confidence filtering step of our PEU pipeline (i.e., only to logits that pass the local-softmax threshold). Token-level scores are then aggregated as in the PEU calculation to produce expert-level utilities.

\begin{table}[H]
    \centering
    \caption{Extended ablation of logit transformations including $e^s$ on accuracy (\%) for the DeepSeek-R1 specialist at 50\% sparsity.}
    \label{tab:logit_filtering_ablation_appendix}
    \small
    \setlength{\tabcolsep}{5pt}
    \begin{tabular}{l l | c c c}
    \toprule
    \textbf{Transformation} & \textbf{Type} & \textbf{MATH-500} & \textbf{GPQA} & \textbf{LCB} \\
    \midrule
    \multicolumn{5}{c}{\textbf{Group 1: No Threshold Filtering}} \\
    \midrule
    $s$ (Raw Logits) & Baseline & 85.20 & 57.07 & 45.59 \\
    $\mathrm{sigmoid}(s)$ & Normalization & 95.20 & 46.97 & 23.90 \\
    $\max(s, \mathrm{sigmoid}(s))$ & Rectifier & 95.00 & 58.08 & 58.09 \\
    \textbf{$e^{s}$} & \textbf{Amplifier} & \textbf{96.80} & \textbf{69.70} & \textbf{64.34} \\
    \midrule
    \multicolumn{5}{c}{\textbf{Group 2: With Threshold Filtering (PreMoE)}} \\
    \midrule
    $s$ (Raw Logits) & Baseline & 97.00 & 68.18 & 67.28 \\
    \textbf{$\mathrm{sigmoid}(s)$} & \textbf{Normalization} & \textbf{97.60} & \textbf{73.23} & \textbf{68.38} \\
    \textbf{$\max(s, \mathrm{sigmoid}(s))$} & \textbf{Rectifier} & \textbf{97.60} & 72.22 & 66.36 \\
    $e^{s}$ & Amplifier & 97.20 & \textbf{73.23} & 66.91 \\
    \bottomrule
    \end{tabular}
\end{table}

\textbf{Observations.} Empirically, without threshold filtering, $e^{s}$ yields the strongest performance by amplifying useful signals in a noisy regime (Table~\ref{tab:logit_filtering_ablation_appendix}, Group 1). When high-confidence filtering is reinstated, the performance of $e^{s}$ converges within a small margin to the top performers (rectifier and/or $\mathrm{sigmoid}(s)$), indicating that extreme amplification is unnecessary once noise is suppressed. For robustness and stability across domains and layers, we therefore adopt $\max(s, \mathrm{sigmoid}(s))$ as the default transformation, and include $e^{s}$ as an additional attempt for completeness.

\section{Additional Experimental Results}
\label{app:additional_results}

This section provides complete numerical results for ablations and analyses discussed in the main text.

\subsection{Component Ablation}
\label{app:component_ablation}

To clearly show the contribution of each component of PreMoE, we conduct an ablation study on DeepSeek-R1 at 50\% sparsity. Table~\ref{tab:component_ablation} progressively adds each component to the baseline.

\begin{table}[H]
    \centering
    \caption{Component ablation on DeepSeek-R1 at 50\% sparsity. Each row adds one component to the previous configuration.}
    \label{tab:component_ablation}
    \small
    \setlength{\tabcolsep}{6pt}
    \begin{tabular}{l|ccc}
    \toprule
    \textbf{Method} & \textbf{MATH-500} & \textbf{GPQA} & \textbf{LCB} \\
    \midrule
    All Logits (baseline) & 3.60 & 28.79 & 0.00 \\
    \midrule
    + TopK filtering (Act-Logits) & 88.20 & 48.48 & 52.94 \\
    \quad + Threshold Filtering & 97.00 & 68.18 & 67.28 \\
    \quad + Logit Transformation & 95.00 & 58.08 & 58.09 \\
    \midrule
    \rowcolor{gray!10} \quad \quad + All (PreMoE) & \textbf{97.60} & \textbf{72.22} & \textbf{66.36} \\
    \bottomrule
    \end{tabular}
\end{table}

\textbf{Key observations:} (1) TopK filtering dramatically improves over using all logits (+84.6 on MATH-500). (2) Threshold filtering alone provides a large boost (+8.8 on MATH-500, +19.7 on GPQA). (3) Logit transformation alone also helps but less than filtering. (4) Combining all components yields the best overall performance, demonstrating their complementary benefits.

\subsection{PEU Ranking Validity: Top vs. Bottom Experts}
\label{app:ranking_validity}

To validate that our PEU rankings meaningfully identify important experts, we compare keeping the top-ranked experts (PreMoE) versus keeping the bottom-ranked experts (Last) on DeepSeek-R1 and openPangu-Ultra.

\begin{table}[H]
    \centering
    \caption{Comparison of keeping top-ranked vs. bottom-ranked experts. DeepSeek-R1 at 50\% sparsity (128/256 experts); openPangu-Ultra at 50\% sparsity (128/256 experts).}
    \label{tab:ranking_validity}
    \small
    \setlength{\tabcolsep}{4pt}
    \begin{tabular}{l|l|ccc}
    \toprule
    \textbf{Model} & \textbf{Method} & \textbf{MATH-500} & \textbf{LCB} & \textbf{GPQA} \\
    \midrule
    \multirow{2}{*}{DeepSeek-R1} & Last-128 (bottom) & 1.4 & 0.0 & 21.72 \\
    & PreMoE (top-128) & \textbf{97.6} & \textbf{66.36} & \textbf{72.22} \\
    \midrule
    \multirow{2}{*}{openPangu-Ultra} & Last-128 (bottom) & 2.2 & 0.0 & 18.69 \\
    & PreMoE (top-128) & \textbf{96.8} & \textbf{66.91} & \textbf{75.76} \\
    \bottomrule
    \end{tabular}
\end{table}

Keeping the bottom-ranked experts results in near-complete collapse across both models (1--2\% on MATH-500, 0\% on LCB), while PreMoE with top-ranked experts achieves near-full accuracy. This stark contrast confirms that PEU rankings accurately identify which experts are critical for task performance.

\subsection{Threshold Strategy Ablation}
\label{app:threshold_ablation}

We compare our adaptive threshold strategy against fixed thresholds on DeepSeek-R1 at 50\% sparsity. Table~\ref{tab:threshold_ablation} shows the results.

\begin{table}[H]
    \centering
    \caption{Ablation of threshold strategies on DeepSeek-R1 at 50\% sparsity.}
    \label{tab:threshold_ablation}
    \small
    \setlength{\tabcolsep}{6pt}
    \begin{tabular}{l|ccc}
    \toprule
    \textbf{Threshold} & \textbf{MATH-500} & \textbf{LCB} & \textbf{GPQA} \\
    \midrule
    Fixed ($r=0.15$) & 94.2 & 64.34 & 68.69 \\
    Fixed ($r=0.3$) & 96.2 & 65.07 & 68.18 \\
    Adaptive (Ours) & \textbf{98.0} & \textbf{64.71} & \textbf{70.71} \\
    \bottomrule
    \end{tabular}
\end{table}

The adaptive threshold consistently outperforms fixed values across all benchmarks. This is because the optimal threshold varies across layers and domains; our layer-wise adaptive rule (Eq.~\ref{eq:adaptive_r}) automatically calibrates to each layer's activation distribution.

\subsection{Local vs. Global Expert Ranking}
\label{app:local_global}

Our default approach profiles the full model once and ranks experts \textit{locally} within each layer, keeping a fixed number (e.g., 96) per layer. To test cumulative error across depth, sequential profiling prunes layers one at a time and re-runs calibration so later layers are scored under the pruned prefix. A second alternative ranks experts \textit{globally} across all layers under the same total budget. Table~\ref{tab:local_global} compares these strategies.

\begin{table}[H]
    \centering
    \caption{Layer-interaction and budget-allocation ablation on DeepSeek-R1 at 62.5\% sparsity (96 experts per layer for local; 5568 total for global).}
    \label{tab:local_global}
    \small
    \setlength{\tabcolsep}{5pt}
    \begin{tabular}{l|ccc}
    \toprule
    \textbf{Strategy} & \textbf{MATH-500} & \textbf{GPQA} & \textbf{LCB} \\
    \midrule
    Random pruning & 48.20 & 32.80 & 18.40 \\
    One-pass local (96/layer) & 96.00 & 64.65 & 61.03 \\
    Sequential multi-pass & 96.40 & \textbf{65.15} & 62.50 \\
    Global budget (5568 total) & \textbf{96.60} & 64.65 & \textbf{63.60} \\
    \bottomrule
    \end{tabular}
\end{table}

\begin{figure}[H]
    \centering
    \includegraphics[width=\columnwidth]{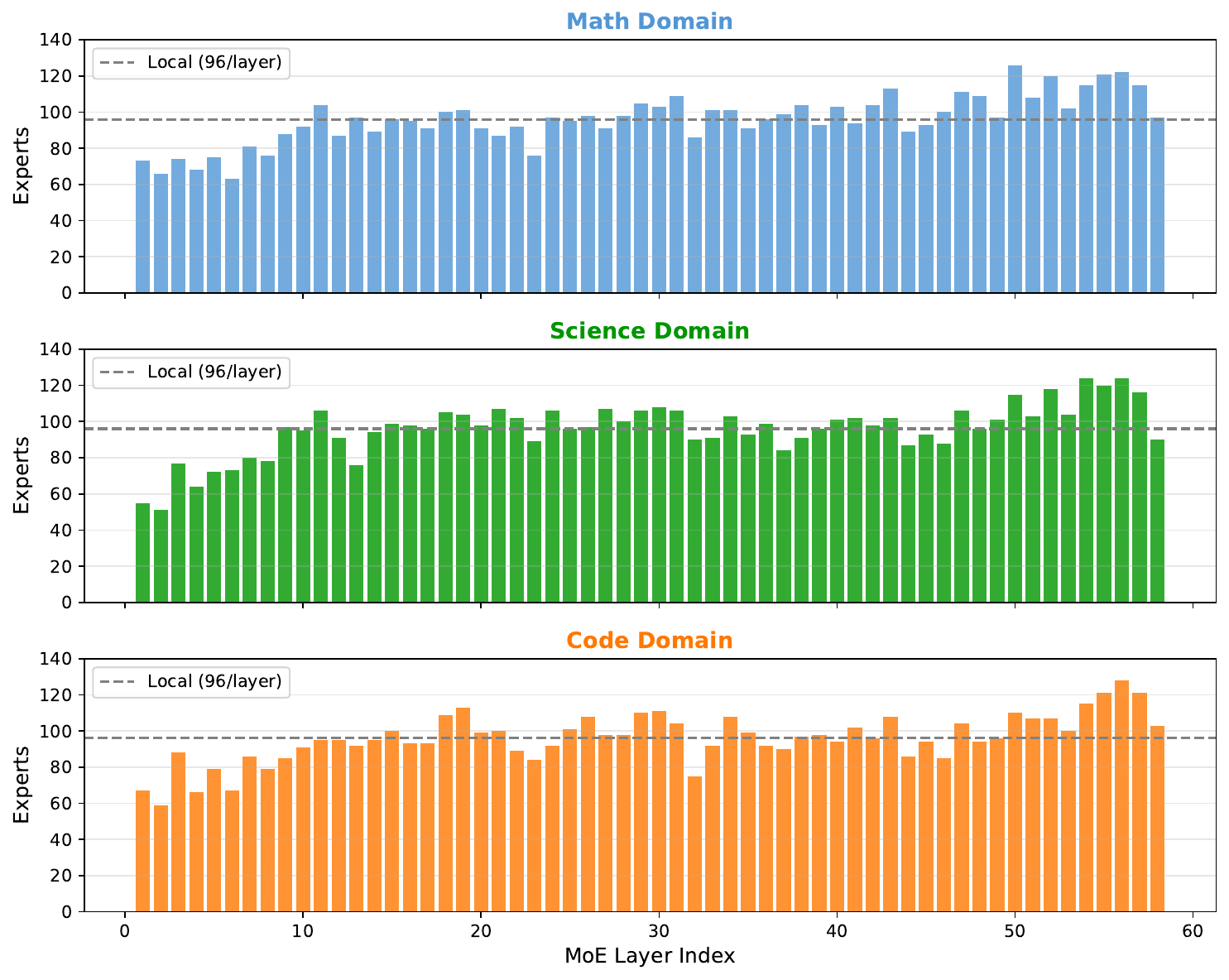}
    \caption{Expert distribution across 58 MoE layers under global ranking for three domains. The dashed line indicates the local baseline (96 experts/layer). Global ranking allocates fewer experts to early layers and more to later layers.}
    \label{fig:global_distribution}
\end{figure}

Sequential profiling improves the three-task average from 73.89 to 74.68, showing that inter-layer effects exist but are modest; the one-pass profile therefore offers a favorable accuracy--compilation-cost trade-off. Global ranking reaches 74.95 and slightly outperforms both local variants, suggesting that some layers benefit from more experts while others need fewer. Figure~\ref{fig:global_distribution} shows that early layers retain fewer experts (51--73), while later layers retain more (up to 128).

\subsection{MMLU-Pro Out-of-Distribution Analysis}
\label{app:mmlu_pro}

We evaluate compiled models on MMLU-Pro, a broad benchmark spanning 14 subject areas, to assess out-of-distribution generalization. Table~\ref{tab:mmlu_pro_breakdown} shows the complete breakdown.

\begin{table}[H]
    \centering
    \caption{MMLU-Pro sub-task breakdown on DeepSeek-R1 at 50\% sparsity. Bold indicates best among compiled models.}
    \label{tab:mmlu_pro_breakdown}
    \footnotesize
    \setlength{\tabcolsep}{2pt}
    \begin{tabular}{l|c|cc|ccc}
    \toprule
    \textbf{Subject} & \textbf{Full} & \textbf{Generalist} & \textbf{+More Data} & \textbf{Math Spec.} & \textbf{Code Spec.} & \textbf{Sci Spec.} \\
    \midrule
    Math & 92.37 & 92.37 & 91.97 & \textbf{92.77} & 89.56 & 88.76 \\
    Physics & 89.72 & 87.85 & \textbf{88.32} & 81.31 & 60.75 & 87.85 \\
    Chemistry & 90.37 & 90.37 & 88.77 & 75.94 & 48.66 & \textbf{89.84} \\
    Law & 67.18 & 35.38 & \textbf{66.67} & 29.23 & 27.69 & 38.97 \\
    Engineering & 81.75 & 73.72 & \textbf{76.64} & 64.96 & 51.09 & 78.10 \\
    Other & 81.06 & 61.36 & \textbf{68.94} & 46.97 & 52.27 & 59.85 \\
    Economics & 84.76 & 76.83 & \textbf{82.32} & 75.00 & 68.90 & 81.10 \\
    Health & 75.51 & 59.86 & \textbf{69.39} & 32.65 & 34.69 & 66.67 \\
    Psychology & 82.17 & 65.89 & \textbf{75.97} & 53.49 & 54.26 & 78.29 \\
    Business & 81.34 & 79.85 & \textbf{82.84} & 76.87 & 78.36 & 78.36 \\
    Biology & 91.18 & 89.22 & \textbf{91.18} & 79.41 & 73.53 & \textbf{91.18} \\
    Philosophy & 78.75 & 63.75 & \textbf{67.50} & 58.75 & 52.50 & 66.25 \\
    Computer & 83.10 & 80.28 & \textbf{83.10} & 73.24 & 74.65 & 67.61 \\
    History & 72.88 & 42.37 & \textbf{52.54} & 38.98 & 27.12 & \textbf{52.54} \\
    \midrule
    \textbf{Average} & 82.29 & 71.36 & \textbf{77.58} & 62.82 & 56.71 & 73.24 \\
    \bottomrule
    \end{tabular}
\end{table}

Key observations: (1) Domain-specific specialists excel in their target areas but degrade sharply elsewhere (e.g., Code Specialist achieves only 27.12\% on History). (2) The base Generalist (calibrated on Math/Code/Science) retains 71.36\% average but struggles on subjects like Law (35.38\%) and History (42.37\%). (3) Augmenting calibration with additional domain data significantly improves generalization to 77.58\%, with the largest gains on Law (+31.29\%), Health (+9.53\%), and Psychology (+10.08\%). This demonstrates that broader calibration coverage directly translates to better out-of-distribution performance.

\subsection{Calibration Context Analysis}
\label{app:calibration_context}

\begin{figure}[H]
    \centering
    \includegraphics[width=0.7\linewidth]{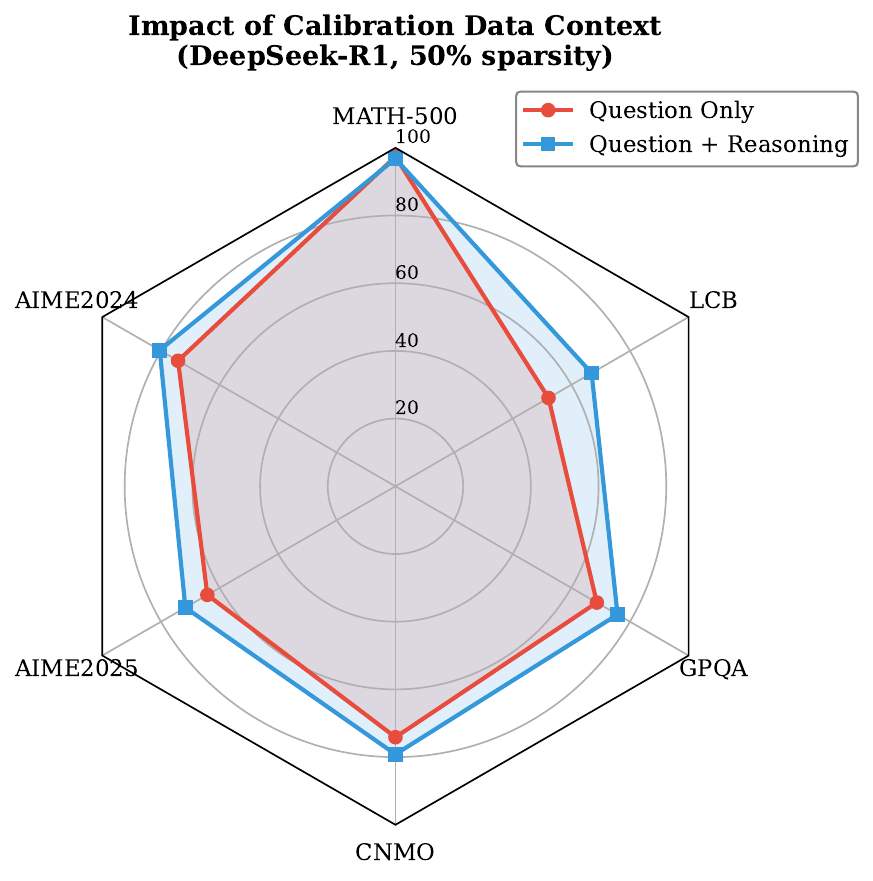}
    \caption{Impact of calibration data context. The radar chart compares performance when extracting computational patterns using \textcolor[RGB]{231,76,60}{question-only} calibration data versus \textcolor[RGB]{52,152,219}{question + reasoning output}. Including the full reasoning context consistently improves accuracy across all six benchmarks, with the largest improvements on AIME 2024 (+6.24\%), LiveCodeBench (+14.70\%), and CNMO (+5.04\%).}
    \label{fig:calibration-context}
\end{figure}

\begin{figure*}[t]
    \centering
    \includegraphics[width=\linewidth]{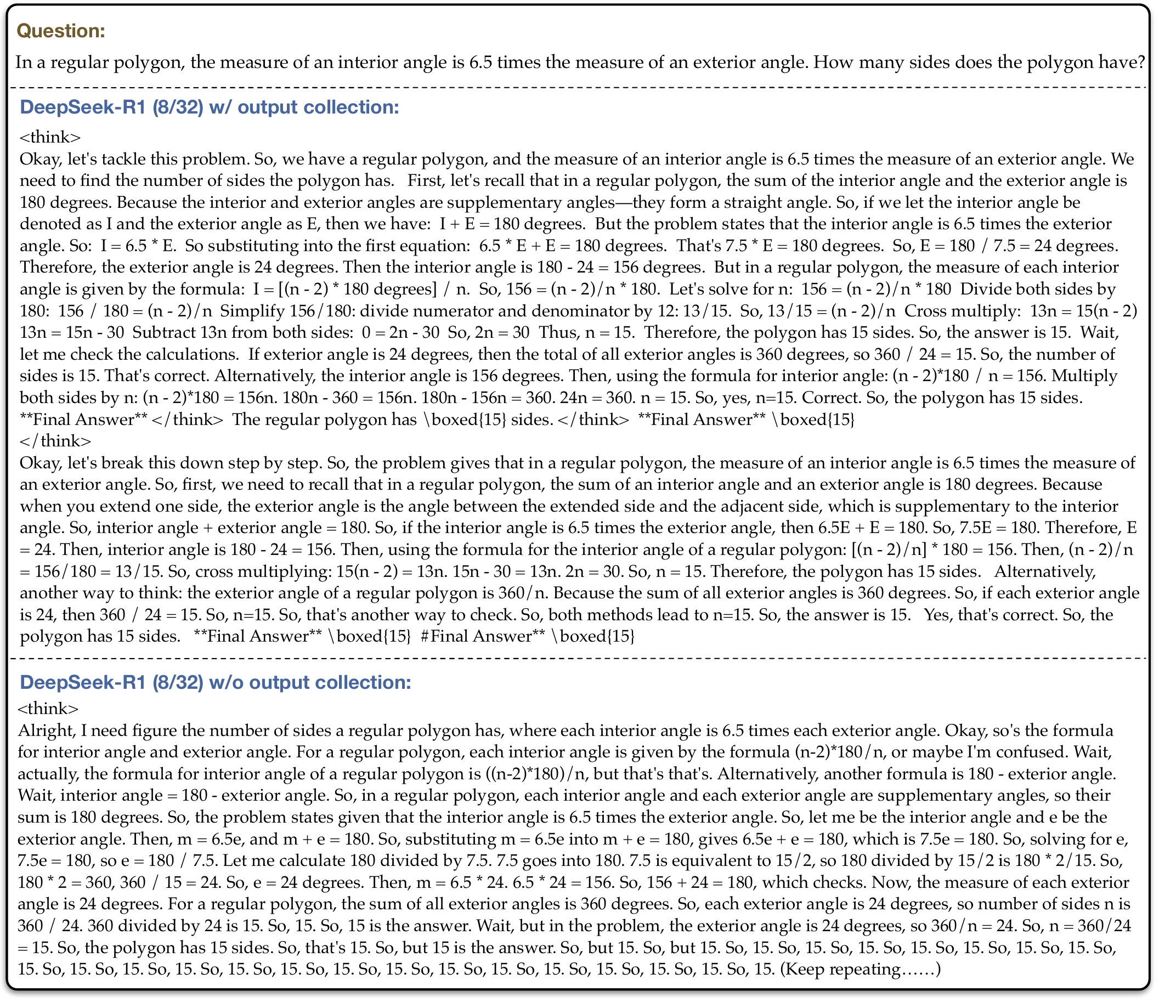} 
    \caption{Comparison of reasoning generation of DeepSeek-R1 when collecting PEU patterns with or without considering the model's reasoning output. The top example uses our default collection strategy (query + reasoning output), while the bottom example derives patterns only from the input query, leading to repetitive output for DeepSeek-R1 (8/32, i.e., 8 experts activated per token out of 32 retained experts).}
    \label{fig:premoe-gen-bad}
\end{figure*}

We find that the quality of the calibration data is paramount. Figure~\ref{fig:calibration-context} shows that including the model's full reasoning output in the calibration context improves accuracy across all benchmarks compared to using only the input problems. Qualitatively, using only inputs can induce repetitive generation loops (see Figure~\ref{fig:premoe-gen-bad} for an extreme case), whereas adding the reasoning outputs stabilizes expert utility patterns and provides richer context for measuring decisive expert activations.

\subsection{Calibration Data Efficiency}
\label{app:calibration_efficiency}

PreMoE requires surprisingly few calibration samples to achieve near-optimal performance. Table~\ref{tab:calibration_efficiency} shows results on DeepSeek-R1 specialists at 50\% sparsity across three domains.

\begin{table}[H]
\centering
\caption{Specialist accuracy vs. calibration sample size on DeepSeek-R1 at 50\% sparsity over multiple random subsets (mean $\pm$ standard deviation). Larger sets mainly reduce variance.}
\label{tab:calibration_efficiency}
\small
\setlength{\tabcolsep}{4pt}
\begin{tabular}{l|cccc}
\toprule
Samples/domain & MATH-500 & GPQA & LCB & Avg \\
\midrule
Full model & 96.60 & 73.23 & 69.12 & 79.65 \\
\midrule
1 & $92.67\pm1.15$ & $42.26\pm2.07$ & $51.84\pm1.46$ & $62.26\pm1.31$ \\
5 & $96.73\pm0.50$ & $70.54\pm1.01$ & $63.97\pm1.23$ & $77.08\pm0.68$ \\
10 & $97.47\pm0.42$ & $70.88\pm0.61$ & $64.83\pm0.74$ & $77.73\pm0.39$ \\
50 & $97.60\pm0.35$ & $71.38\pm0.51$ & $66.30\pm0.56$ & $78.43\pm0.32$ \\
100 & $97.73\pm0.31$ & $71.72\pm0.44$ & $66.42\pm0.48$ & $78.62\pm0.25$ \\
Default & 97.60 & 72.22 & 66.36 & 78.73 \\
\bottomrule
\end{tabular}
\end{table}

Performance stabilizes rapidly for specialists: just 5--10 samples approach the full accuracy. For generalists, using only 10 samples per domain (30 total) also matches full performance (Table~\ref{tab:calibration_efficiency_generalist}).

\begin{table}[H]
\centering
\caption{Generalist accuracy with reduced calibration data on DeepSeek-R1 at 50\% sparsity.}
\label{tab:calibration_efficiency_generalist}
\small
\setlength{\tabcolsep}{3pt}
\begin{tabular}{l|cccccc|c}
\toprule
Calibration & MATH-500 & GPQA & LCB & AIME 24 & AIME 25 & CNMO & Avg \\
\midrule
Full & 96.6 & 73.23 & 69.12 & 77.08 & 65.83 & 71.18 & 75.52 \\
10+10+10 (30 total) & 96.4 & 73.74 & 68.38 & 79.58 & 62.92 & 74.88 & 75.98 \\
\bottomrule
\end{tabular}
\end{table}

\subsection{Accuracy at Various Sparsity Levels}

\begin{table}[H]
    \centering
    \caption{DeepSeek-R1 Domain-Specific Specialists: accuracy (\%) vs. sparsity. Average is over MATH-500, GPQA, LCB.}
    \label{tab:sparsity_specialist}
    \small
    \setlength{\tabcolsep}{5pt}
    \begin{tabular}{l | c | c c c | c}
    \toprule
    \textbf{\# Experts} & \textbf{Sparsity} & \textbf{MATH-500} & \textbf{GPQA} & \textbf{LCB} & \textbf{Avg} \\
    \midrule
    256 & 0\% & 96.60 & 73.23 & 69.12 & 79.65 \\
    128 & 50\% & 97.60 & 72.22 & 66.36 & 78.73 \\
    96 & 62.5\% & 96.00 & 64.65 & 61.03 & 73.89 \\
    64 & 75\% & 93.40 & 45.45 & 49.63 & 62.83 \\
    \bottomrule
    \end{tabular}
\end{table}

\begin{table}[H]
    \centering
    \caption{DeepSeek-R1 High-Efficiency Generalist: accuracy (\%) vs. sparsity. Average is over MATH-500, GPQA, LCB.}
    \label{tab:sparsity_generalist}
    \small
    \setlength{\tabcolsep}{5pt}
    \begin{tabular}{l | c | c c c | c}
    \toprule
    \textbf{\# Experts} & \textbf{Sparsity} & \textbf{MATH-500} & \textbf{GPQA} & \textbf{LCB} & \textbf{Avg} \\
    \midrule
    256 & 0\% & 96.60 & 73.23 & 69.12 & 79.65 \\
    128 & 50\% & 97.40 & 67.68 & 68.01 & 77.69 \\
    96 & 62.5\% & 93.60 & 62.63 & 56.99 & 71.07 \\
    64 & 75\% & 89.80 & 38.38 & 20.22 & 49.46 \\
    \bottomrule
    \end{tabular}
\end{table}

\paragraph{Specialist vs. Generalist Degradation Under Extreme Sparsity.}
Comparing Tables~\ref{tab:sparsity_specialist} and \ref{tab:sparsity_generalist}, we observe that domain-specific specialists degrade more gracefully at extreme sparsity than multi-domain generalists. At 75\% sparsity, the specialist average (62.83\%) significantly outperforms the generalist (49.46\%). The gap is particularly pronounced on GPQA (45.45\% vs 38.38\%) and especially LCB (49.63\% vs 20.22\%). This indicates that when expert budgets are severely constrained, dedicating the limited experts to a single domain yields better task-specific performance than attempting multi-domain coverage. This finding suggests a trade-off between model versatility and efficiency: at moderate sparsity (50\%), the generalist retains broad capabilities with minimal loss, but at extreme sparsity (75\%), domain-focused specialists become the more practical choice.

\subsection{Complete Deployment Efficiency Metrics}

\begin{table}[H]
    \centering
    \caption{Complete deployment efficiency metrics for DeepSeek-R1 at various sparsity levels. Size refers to the number of model parameters in billions (B). All measurements are taken on servers with Ascend 910B2-64GB NPUs.}
    \label{tab:efficiency_metrics}
    \footnotesize
    \setlength{\tabcolsep}{3pt}
    \begin{tabular}{c c c c c c}
    \toprule
    \textbf{Sparsity} & \textbf{\# Experts} & \textbf{\# NPUs} & \textbf{Latency (ms/tok)} & \textbf{Size (B)} & \textbf{Throughput (tok/s)} \\
    \midrule
    0\% & 256 & 64 & 115.35 & 670.92 & 52.01 \\
    50\% & 128 & 32 & 93.72 & 343.96 & 64.02 \\
    75\% & 64 & 16 & 73.20 & 180.49 & 81.97 \\
    \midrule
    \multicolumn{6}{l}{\textbf{Relative Improvement (vs. 0\% sparsity)}} \\
    \midrule
    50\% & -- & 2× fewer & 1.23× faster & 1.95× smaller & 1.23× higher \\
    75\% & -- & 4× fewer & 1.58× faster & 3.72× smaller & 1.58× higher \\
    \bottomrule
    \end{tabular}
\end{table}

These metrics demonstrate that PreMoE enables significant infrastructure savings while maintaining high accuracy. At 50\% sparsity, we halve the deployment cost (NPUs and parameters) while improving inference speed by 23\%. At 75\% sparsity, we achieve 4× NPU reduction and 58\% throughput improvement, though with accuracy trade-offs on some benchmarks as shown in the main text.

\section{Additional Robustness, Cost, and Scope Analyses}
\label{app:camera_ready_analyses}

\subsection{Calibration Source and Token-Context Robustness}
\label{app:calibration_robustness}

We isolate two calibration factors on DeepSeek-R1 at 50\% sparsity. Source robustness uses domain-aligned prompts from different datasets without evaluation-benchmark selection. Token-context robustness holds the source fixed and varies which parts of the full-model trace are used to collect router statistics. Table~\ref{tab:calibration_source_context_full} reports all six benchmarks.

\FloatBarrier
\begin{table}[H]
\centering
\caption{Calibration source and token-context sensitivity on DeepSeek-R1 specialists at 50\% sparsity. All design choices are fixed across benchmarks.}
\label{tab:calibration_source_context_full}
\small
\setlength{\tabcolsep}{2pt}
\begin{tabular}{ll|cccccc|c}
\toprule
\textbf{Factor} & \textbf{Setting} & \textbf{MATH-500} & \textbf{AIME 24} & \textbf{AIME 25} & \textbf{CNMO 24} & \textbf{GPQA} & \textbf{LCB} & \textbf{Avg} \\
\midrule
\multirow{3}{*}{Source}
& Default & 97.60 & 79.58 & 68.33 & 75.00 & 72.22 & 66.36 & 76.52 \\
& Mixed source & 97.20 & 78.75 & 67.92 & 74.65 & 71.72 & 65.99 & 76.04 \\
& Alternative only & 96.80 & 78.33 & 67.50 & 74.31 & 71.21 & 65.44 & 75.60 \\
\midrule
\multirow{3}{*}{Tokens}
& Question only & 92.80 & 65.42 & 54.58 & 61.81 & 63.13 & 55.88 & 65.60 \\
& Generated reasoning only & 96.60 & 77.50 & 67.08 & 73.96 & 70.71 & 64.52 & 75.06 \\
& Question + reasoning & 97.60 & 79.58 & 68.33 & 75.00 & 72.22 & 66.36 & \textbf{76.52} \\
\bottomrule
\end{tabular}
\end{table}

Alternative sources remain within 0.92 average points of the default, so PreMoE is not tied to one dataset. Token context has a larger effect: question-only calibration observes prefill routing but omits the longer decoding trajectory where most reasoning computation occurs. This architectural distinction motivates our globally fixed question-plus-reasoning protocol.

\subsection{Generalist Mixture Sensitivity}
\label{app:mixture_sensitivity}

The generalist definition in Appendix~\ref{app:method_details_2} permits deployment weights $w_d$. Table~\ref{tab:mixture_sensitivity} varies the Math:Science:Code ratio while keeping the expert budget fixed. The trade-offs are smooth and interpretable: math-heavy profiles improve math and reduce code, while code-heavy profiles do the reverse. Equal weighting is the best deployment-agnostic average, but the method supports non-uniform workload priorities without changing the compilation algorithm.

\FloatBarrier
\begin{table}[H]
\centering
\caption{Generalist mixture-ratio sensitivity on DeepSeek-R1 at 50\% sparsity.}
\label{tab:mixture_sensitivity}
\small
\setlength{\tabcolsep}{4pt}
\begin{tabular}{l|cccccc|c}
\toprule
\textbf{Math:Sci:Code} & \textbf{MATH-500} & \textbf{AIME 24} & \textbf{AIME 25} & \textbf{CNMO 24} & \textbf{GPQA} & \textbf{LCB} & \textbf{Avg} \\
\midrule
Full model & 96.60 & 77.08 & 65.83 & 71.18 & 73.23 & 69.12 & 75.51 \\
\midrule
1:1:1 (default) & 96.40 & 78.33 & 70.42 & 75.17 & 70.71 & 65.07 & \textbf{76.02} \\
2:1:1 (math-heavy) & 97.20 & 79.17 & 70.83 & 75.69 & 69.70 & 62.87 & 75.91 \\
1:2:1 (science-heavy) & 95.80 & 76.67 & 68.75 & 73.96 & 72.73 & 63.60 & 75.25 \\
1:1:2 (code-heavy) & 95.60 & 76.25 & 68.33 & 73.61 & 69.70 & 67.65 & 75.19 \\
2:2:1 (reasoning-heavy) & 96.80 & 78.75 & 70.00 & 75.35 & 71.72 & 63.24 & 75.98 \\
1:2:2 (non-math-heavy) & 95.40 & 75.83 & 68.33 & 73.44 & 72.22 & 67.28 & 75.42 \\
\bottomrule
\end{tabular}
\end{table}

\subsection{Held-Out Validation and Decontamination}
\label{app:calibration_validation}

To check that calibration format was not selected on the reported test suite, we construct a small validation set from GSM8K and MBPP, neither of which appears among our evaluation benchmarks. The ranking in Table~\ref{tab:heldout_format} matches the main evaluation: question plus generated reasoning performs best. We treat this as a sanity check rather than evidence that these two datasets cover every deployment distribution.

\FloatBarrier
\begin{table}[H]
\centering
\caption{Calibration-format sanity check on held-out validation tasks.}
\label{tab:heldout_format}
\small
\setlength{\tabcolsep}{5pt}
\begin{tabular}{lccc}
\toprule
\textbf{Calibration format} & \textbf{GSM8K} & \textbf{MBPP} & \textbf{Avg} \\
\midrule
Question only & 88.32 & 54.20 & 71.26 \\
Generated reasoning only & 94.16 & 63.40 & 78.78 \\
Question + reasoning & \textbf{95.07} & \textbf{65.80} & \textbf{80.44} \\
\bottomrule
\end{tabular}
\end{table}

We also compare every normalized calibration prompt with every evaluation prompt using (i) 13-gram Jaccard similarity, flagging scores above 0.8, and (ii) \texttt{all-MiniLM-L6-v2} embedding cosine similarity, flagging scores above 0.9. Table~\ref{tab:near_duplicate_check} shows no flagged pair. Three lower-threshold pairs in $[0.85,0.9]$ are generic templates with different instances; conservatively removing them changes the DeepSeek-R1 specialist average from 76.52 to 76.47. We will release calibration identifiers and the checking scripts. More importantly, calibration only aggregates router logits: it does not update weights, use ground-truth answers, or fit benchmark outputs.

\FloatBarrier
\begin{table}[H]
\centering
\caption{Near-duplicate checks between calibration and evaluation prompts.}
\label{tab:near_duplicate_check}
\small
\setlength{\tabcolsep}{4pt}
\begin{tabular}{lrrrr}
\toprule
\textbf{Domain} & \textbf{Samples} & \textbf{13-gram $>$.8} & \textbf{Embed $>$.9} & \textbf{Embed $>$.85} \\
\midrule
Math & 800 & 0 & 0 & 2 \\
Science & 200 & 0 & 0 & 0 \\
Code & 600 & 0 & 0 & 1 \\
Dialogue & 600 & 0 & 0 & 0 \\
\bottomrule
\end{tabular}
\end{table}

\subsection{One-Time Profiling Cost}
\label{app:offline_cost}

PreMoE is training-free, not calibration-free. Table~\ref{tab:offline_cost} reports the one-time profiling cost and an illustrative break-even point assuming that the compiled instance costs approximately half as much per deployment token as the parent model. Only top-routed expert identifiers and scalar router statistics are accumulated; full-vocabulary logits and expert activations are not stored. The default profile amortizes after a few million deployment tokens, and Table~\ref{tab:calibration_efficiency} shows that a lower-cost 5--10-sample profile already recovers most of the accuracy.

\FloatBarrier
\begin{table}[H]
\centering
\caption{One-time offline profiling cost. The three-domain row excludes the optional dialogue augmentation so its sample count equals the sum of the three specialist sets.}
\label{tab:offline_cost}
\small
\setlength{\tabcolsep}{6pt}
\begin{tabular}{lrrrr}
\toprule
\textbf{Profile} & \textbf{Samples} & \textbf{Tokens} & \textbf{Time (accelerator-h)} & \textbf{Break-even tokens} \\
\midrule
Math specialist & 800 & 1.2M & 0.42 & 2.4M \\
Science specialist & 200 & 0.3M & 0.11 & 0.6M \\
Code specialist & 600 & 0.9M & 0.31 & 1.8M \\
Three-domain generalist & 1600 & 2.4M & 0.84 & 4.8M \\
\bottomrule
\end{tabular}
\end{table}

\subsection{Static Profiles and Profile Libraries}
\label{app:profile_library}

A compiled expert mask is static during inference. To quantify the cost of using one mask for heterogeneous traffic, Table~\ref{tab:profile_library} applies each narrow profile to all three tasks, compares a mixed generalist, and reports an oracle that selects the matching specialist for each known domain. Narrow profiles transfer imperfectly, the generalist is a better single-mask default, and the oracle shows the potential of a profile library. The oracle assumes domain labels and is not an evaluated online selector.

\FloatBarrier
\begin{table}[H]
\centering
\caption{A single static mask versus a domain-profile library on DeepSeek-R1 at 50\% sparsity.}
\label{tab:profile_library}
\small
\setlength{\tabcolsep}{4pt}
\begin{tabular}{lrrrr}
\toprule
\textbf{Inference profile} & \textbf{MATH-500} & \textbf{GPQA} & \textbf{LCB} & \textbf{Avg} \\
\midrule
Math-specific & 97.4 & 67.7 & 60.7 & 75.3 \\
Science-specific & 94.8 & 73.2 & 62.1 & 76.7 \\
Code-specific & 94.2 & 68.7 & 68.4 & 77.1 \\
Generalist (1:1:1) & 96.4 & 70.7 & 65.1 & 77.4 \\
Oracle domain profile & \textbf{97.4} & \textbf{73.2} & \textbf{68.4} & \textbf{79.7} \\
\bottomrule
\end{tabular}
\end{table}

\subsection{Additional Comparison with REAP}
\label{app:reap_comparison}

REAP~\citep{lasby2025reap} scores an expert using its router gate multiplied by its output norm, whereas PreMoE uses confidence-filtered router logits. Table~\ref{tab:reap_quality_full} uses the same DeepSeek-R1 backbone, calibration data, 50\% layer-wise budget, and evaluation protocol. Scores are normalized to the dense model to isolate retention under pruning and should not be mixed with the absolute results in Table~\ref{tab:main_specialist_results}.

\FloatBarrier
\begin{table}[H]
\centering
\caption{Normalized pruning quality under the matched REAP comparison. EAN denotes expert activation norm.}
\label{tab:reap_quality_full}
\small
\setlength{\tabcolsep}{5pt}
\begin{tabular}{lcc|rrrr}
\toprule
\textbf{Method} & \textbf{Router logits} & \textbf{Activation norms} & \textbf{MATH-500} & \textbf{GPQA} & \textbf{LCB} & \textbf{Avg} \\
\midrule
Dense/original & -- & -- & 100.0 & 100.0 & 100.0 & 100.0 \\
Frequency & No & No & 86.5 & 84.0 & 78.0 & 82.8 \\
EAN & No & Yes & 91.5 & 89.0 & 85.5 & 88.7 \\
REAP & Yes & Yes & 94.0 & 92.5 & 89.5 & 92.0 \\
PreMoE & Yes & No & 93.2 & 91.8 & 88.7 & 91.2 \\
PreMoE + ActNorm & Yes & Yes & \textbf{94.3} & \textbf{92.7} & \textbf{90.0} & \textbf{92.3} \\
\bottomrule
\end{tabular}
\end{table}

Activation norms improve PreMoE by 1.1 normalized points, but are not required to approach REAP. The practical distinction is the extra signal collected on top of the shared sparse-MoE forward pass. Both methods can accumulate $O(LE)$ final scalars online, but REAP must reduce a $d$-dimensional expert output for every routed token-expert pair, while PreMoE processes scalar router scores. On approximately $4.2\times10^5$ calibration tokens with $d=7168$, our implementation measures about 8\,MB versus 1.6\,GB peak extra device memory and 12\% versus 75\% wall-clock overhead. Thus, the claim is lower activation-level profiling overhead, not avoidance of the common calibration forward pass or asymptotically smaller persistent statistics.

\subsection{Router-Logit-Free Settings}
\label{app:routerless}

Full PreMoE targets learned-router MoEs. Fixed hash routing~\citep{roller2021hash} provides no preference logits, so PEU cannot be applied unchanged. Table~\ref{tab:routerless} removes access to logit magnitudes and evaluates increasingly informative activation-side substitutes. Rank-1 counts and hidden-state utility recover part of the loss, but still trail full PEU substantially. Extending proactive compilation to router-logit-free architectures is therefore a separate utility-estimation problem rather than a drop-in use of PreMoE.

\FloatBarrier
\begin{table}[H]
\centering
\caption{Router-logit-free utility signals on DeepSeek-R1 at 50\% sparsity. The ablation emulates the information available without learned-router preference magnitudes.}
\label{tab:routerless}
\small
\setlength{\tabcolsep}{2pt}
\begin{tabular}{lclrrrr}
\toprule
\textbf{Method} & \textbf{Uses logits?} & \textbf{Utility signal} & \textbf{MATH-500} & \textbf{GPQA} & \textbf{LCB} & \textbf{Avg} \\
\midrule
Frequency pruning & No & Activation count & 88.80 & 33.33 & 5.88 & 42.67 \\
Rank-1 frequency & No & Decisive activation count & 94.20 & 58.50 & 52.00 & 68.23 \\
Logit-free extension & No & Activation + hidden-state utility & 94.80 & 60.10 & 55.30 & 70.07 \\
Full PreMoE & Yes & PEU from router logits & \textbf{97.60} & \textbf{72.22} & \textbf{66.36} & \textbf{78.73} \\
\bottomrule
\end{tabular}
\end{table}

\subsection{Contextual Comparison at Similar Served Size}
\label{app:served_size_context}

For deployment context, Table~\ref{tab:served_size_context} evaluates Qwen3-235B-A22B and PreMoE-pruned DeepSeek-R1 with the same harness, prompt templates, and decoding setup. ``Served parameters'' means weights that must be stored for the instance, not parameters activated per token. This comparison is not controlled for pretraining, post-training, architecture, or reasoning recipe and does not establish that pruning a large MoE dominates training a native smaller model. The controlled evidence for PreMoE remains the same-parent comparison in Table~\ref{tab:main_specialist_results}.

\FloatBarrier
\begin{table}[H]
\centering
\caption{Additional matched-harness context at similar served parameter counts.}
\label{tab:served_size_context}
\small
\setlength{\tabcolsep}{4pt}
\begin{tabular}{lrrrr}
\toprule
\textbf{Model} & \textbf{Served params} & \textbf{MATH-500} & \textbf{AIME 24} & \textbf{LCB} \\
\midrule
DeepSeek-R1 full & 671B & 96.6 & 77.1 & 69.1 \\
PreMoE 50\% & $\sim$336B & 97.4 & 78.3 & 65.4 \\
PreMoE 70\% & $\sim$202B & 95.2 & 73.8 & 60.9 \\
Qwen3-235B-A22B & 235B & 97.1 & \textbf{82.6} & \textbf{68.4} \\
\bottomrule
\end{tabular}
\end{table}

The 50\%-sparse instance retains the parent model closely but remains larger than Qwen3-235B-A22B; the 70\%-sparse instance is closer in served size and trails Qwen3 on AIME and code. We therefore position PreMoE as a deployment-time option for adapting an existing parent model, not as a substitute for a native model trained at the target size.

\section{Discussions, Limitations, and Future Work}
\label{appendix:discuss_limitations}

\textbf{Discussions.}
PreMoE demonstrates that MoE models across scales (30B--718B parameters) can be efficiently deployed through proactive compilation, achieving up to 50\% sparsity with minimal performance loss. Our experiments show that the approach is remarkably data-efficient: just 5--10 calibration samples achieve near-optimal performance (Appendix~\ref{app:calibration_efficiency}), and including reasoning context significantly improves pattern quality (Appendix~\ref{app:calibration_context}). The cross-domain expert overlap analysis reveals that high-utility experts are highly domain-specific while lower-ranked experts are more general, validating our synthesis approach for multi-domain generalists. This work establishes computational patterns as a practical tool for model specialization, enabling significant infrastructure savings while maintaining model capabilities.

\textbf{Limitations and Future Work.}
PreMoE remains a static, calibration-dependent compiler. Its strong source and seed robustness does not eliminate degradation when deployment contains domains absent from calibration; MMLU-Pro and profile-library results make this boundary explicit. Automatic shift detection and online profile selection are not evaluated. The safe pruning ratio is also model- and workload-dependent, and our architectures contain only 128 or 256 routed experts per layer. Moreover, PEU assumes a learned router with informative logits; fixed hash-routing models~\citep{roller2021hash} require a different utility signal (Appendix~\ref{app:routerless}). Finally, one-pass scoring ignores small inter-layer effects, and router preference does not directly measure functional similarity between expert transformations. Sequential profiling and activation norms recover modest gains, while weight-space or Jacobian-alignment signals~\citep{lowell2024alignment} are promising complementary criteria. These extensions trade PreMoE's low profiling cost for additional information.

\section{Broader Literature}
\label{app:broader_literature}

\subsection{Mixture-of-Experts Architectures}
The Mixture-of-Experts (MoE) paradigm has been instrumental in scaling neural networks to trillions of parameters while keeping computational costs manageable~\citep{shazeer2017outrageously, lepikhin2020gshard}. By routing each input token to a small subset of ``expert'' sub-networks, MoE models such as GShard, Switch Transformers~\citep{fedus2022switch}, and more recent large language models like Mixtral~\citep{jiang2024mixtral} and DeepSeek-R1~\citep{guo2025deepseek} achieve performance comparable to much larger dense models through dynamic, sparse activation of parameters.

\subsection{Efficiency in Large Language Models}
The challenge of deploying massive LLMs has spurred extensive research into model efficiency. Techniques such as quantization~\citep{dettmers2022gpt3, frantar2022gptq}, which reduces the numerical precision of model weights, and knowledge distillation~\citep{hinton2015distilling}, where a smaller student model is trained to mimic a larger teacher, are commonplace. Structural pruning~\citep{frankle2018lottery, hoefler2021sparsity, pei2024fusegpt, pei2025cmoe} aims to remove entire neurons or weights from dense models. While effective, these methods are often applied globally and do not typically account for the specific, input-dependent activation patterns unique to MoE architectures.

\section{Examples of Generations}
\label{appendix:example}
In this section, we present additional examples of the reasoning generation capabilities of the DeepSeek-R1 model under various pruning ratios facilitated by PreMoE.


\begin{figure*}[t]
    \centering
    \includegraphics[width=\linewidth]{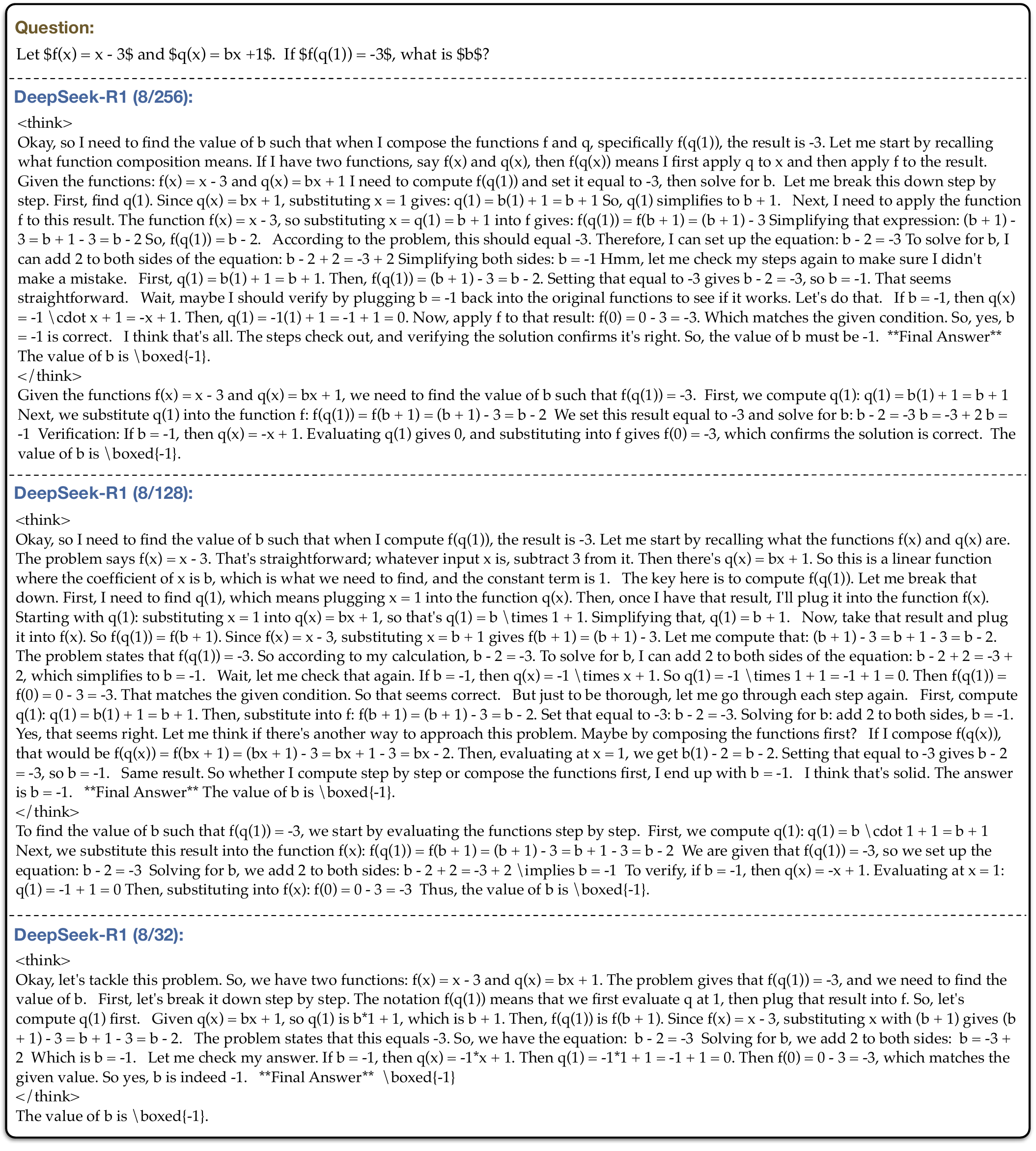} 
    \caption{Example 1 of reasoning generation of DeepSeek-R1 with different sparsity levels using PreMoE. Notation: activated/retained experts (e.g., 8/32 means 8 experts activated per token out of 32 retained).}
    \label{fig:premoe-gen1}
\end{figure*}

\begin{figure*}[t]
    \centering
    \includegraphics[width=\linewidth]{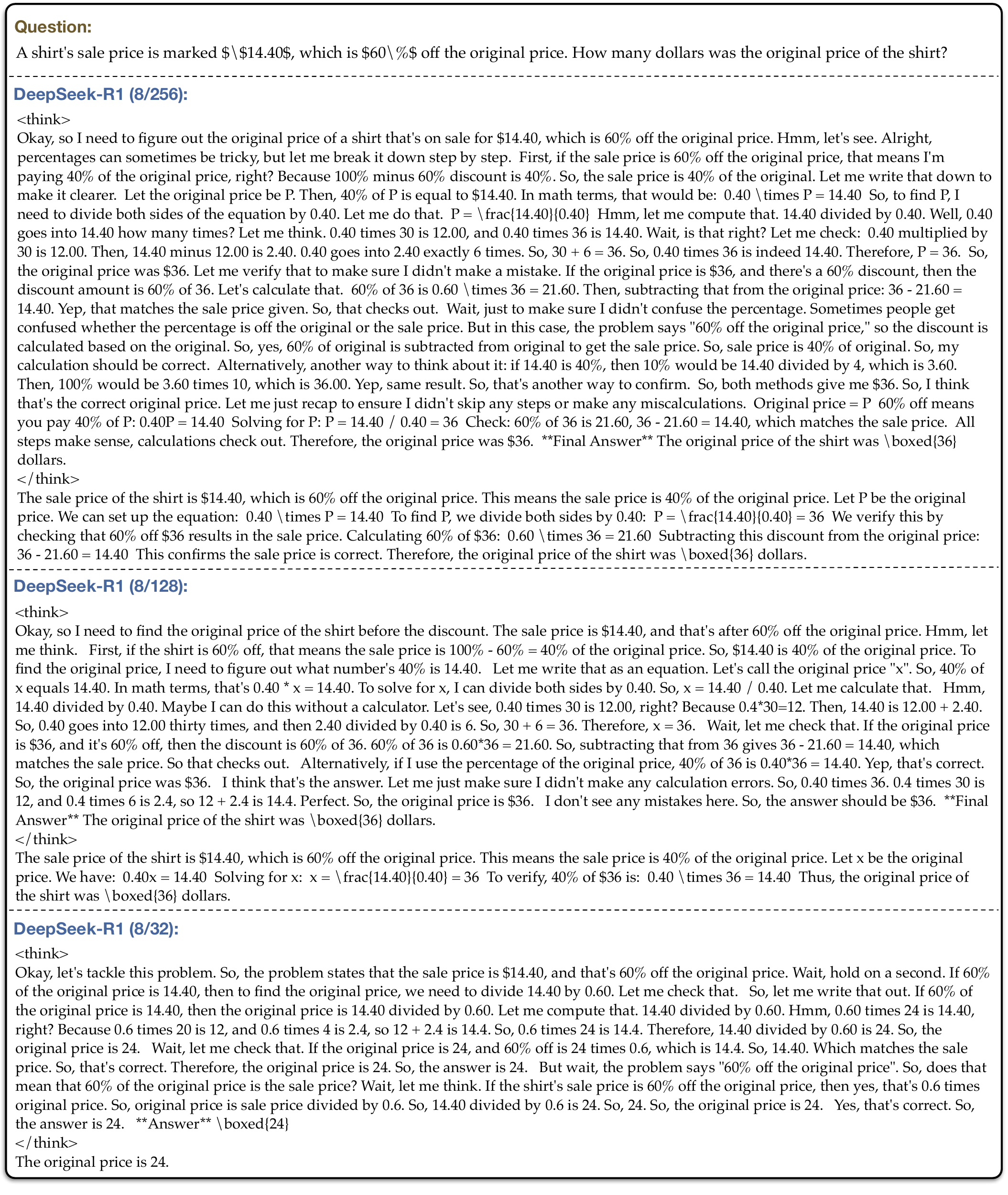} 
    \caption{Example 2 of reasoning generation of DeepSeek-R1 with different sparsity levels using PreMoE. Notation: activated/retained experts.}
    \label{fig:premoe-gen2}
\end{figure*}

\end{document}